\documentclass{article}

\usepackage[english]{babel}

\usepackage{microtype}
\usepackage{graphicx}
\usepackage{subfig}
\usepackage{booktabs} 
\usepackage{bbm}
\usepackage[inline]{enumitem}
\setlist[enumerate]{label=\roman*)}
\usepackage{hyperref}

\newcommand{\SC}{{\cal S}}
\newcommand{\A}{{\cal A}}

\usepackage[accepted]{icml2023}

\usepackage{amsmath}
\usepackage{amssymb}
\usepackage{mathtools}
\usepackage{amsthm}

\usepackage[capitalize,noabbrev]{cleveref}

\theoremstyle{plain}
\newtheorem{theorem}{Theorem}[section]
\newtheorem{proposition}[theorem]{Proposition}
\newtheorem{lemma}[theorem]{Lemma}

\theoremstyle{definition}

\theoremstyle{remark}

\usepackage[textsize=tiny]{todonotes}

\icmltitlerunning{Limits of Actor-Critic Algorithms for Decision Tree Policies Learning in IBMDPs}

\begin{document}

\twocolumn[
\icmltitle{Limits of Actor-Critic Algorithms for Decision Tree Policies Learning in IBMDPs}



\icmlsetsymbol{equal}{*}

\begin{icmlauthorlist}
\icmlauthor{Hector Kohler}{yyy}
\icmlauthor{Riad Akrour}{yyy}
\icmlauthor{Philippe Preux}{yyy}
\end{icmlauthorlist}

\icmlaffiliation{yyy}{Université de Lille, CNRS, Inria, CRIStAL, Centrale - Lille, France}

\icmlcorrespondingauthor{Hector Kohler}{hector.kohler@inria.fr}

\icmlkeywords{Machine Learning, ICML}

\vskip 0.3in
]



\printAffiliationsAndNotice{\icmlEqualContribution} 
\begin{abstract}
Interpretability of AI models allows for user safety checks to build trust in such AIs. In particular, Decision Trees (DTs) provide a global look at the learned model and transparently reveal which features of the input are critical for making a decision. However, interpretability is hindered if the DT is too large. To learn compact trees, a recent Reinforcement Learning (RL) framework has been proposed to explore the space of DTs using deep RL. This framework augments a decision problem (e.g. a supervised classification task) 
with additional actions that gather information about the features of an otherwise hidden input. By appropriately penalizing these actions, the agent learns to optimally trade-off size and performance of DTs. In practice, a reactive policy for a partially observable Markov decision process (MDP) needs to be learned, which is still an open problem. We show in this paper that deep RL can fail even on simple toy tasks of this class. However, when the underlying decision problem is a supervised classification task, we show that finding the optimal tree can be cast as a fully observable Markov decision problem and be solved efficiently, giving rise to a new family of algorithms for learning DTs that go beyond the classical greedy maximization ones.   
\end{abstract}

\section{Introduction}
The last decade or so has seen a surge in the performance of machine learning models, whether in supervised learning~\cite{AlexNet} or Reinforcement Learning ~\citet{mnih2015human}. These achievements rely on deep neural models that are often described as black-box~\cite{Murdoch, Guidotti18, Arrieta}, trading interpretability for performance. In many real world tasks, predictive models can hide undesirable biases~(see e.g. Sec. 2 in \citet{Guidotti18} for a list of such occurrences) hindering trustworthiness towards AIs. Gaining trust is one of the primary goals of interpretability~(see Sec.\@ 2.4 of \citet{Arrieta} for a literature review) along with informativeness requests, i.e.\@ the ability for a model to provide information on why a given decision was taken. The computational complexity of such informativeness requests can be measured objectively, and \citet{Barcelo20} showed that multi-layer neural networks cannot answer these requests in polynomial time, whereas several of those are in polynomial time for linear models and decision trees.

In contrast to deep neural models, DTs provide a global look at the learned model and transparently reveal which features of the input are used in taking a particular decision. This is referred to as global \cite{Guidotti18} or model-based \cite{Murdoch} interpretability, as opposed to post-hoc interpretability \cite{Murdoch, Arrieta}. Even though DTs are globally intepretable, they have also been used in prior work for post-hoc interpretability of deep neural models, e.g.\@ in image classification \cite{Zhang19} or RL \cite{Viper}. The latter work provides another motivation for DTs, as their simpler nature allowed to make a stability analysis of the resulting controllers and provided theoretical guarantees of their efficacy. In the 54 papers reviewed in \citet{Guidotti18}, over $25\%$ use DTs as the interpretable model and over $50\%$ the more general class of decision rules\footnote{While we focus on learning decision trees, the proposed RL algorithms uses very sparingly the nature of the underlying interpretable model, making it easier to extend to broader classes of interpretable models in the future. See Sec.\@ \ref{sec:disc}.}.

DTs are a common interpretable model and it is thus important to improve 
their associated learning algorithms. However, interpretability of DTs is hindered if the tree grows too large. The quantification of what is too large might vary greatly depending on the desired type of simulability, that is whether we want individual paths from root to leaf to be short or the total size of the tree to be small (\citet[p.\@ 13]{Lipton}). In both cases, an algorithmic mechanism to control these tree metrics and to manage the inevitable trade-off between interpretability and performance is necessary. One of the main challenges for learning DTs is that it is a discrete optimization problem that cannot, a priori, be solved via gradient descent. Algorithms such as CART \cite{Cart} build a DT by greedily maximizing the information gain---a performance related criteria. Interpretability can then be controlled by fixing a maximal tree depth, or by using post-processing pruning algorithms \cite{Bradford, Prodromidis}. Unfortunately, this two-step process provides no guaranty that the resulting DT is achieving an optimal interpretability-performance trade-off.

An alternative way to learn DTs, that inherently takes into account the interpretability-performance trade-off, is the recently proposed framework of Iterative Bounding Markov Decision Processes (IBMDPs, ~\cite{IBMDP}). An IBMDP extends a base MDP state space with \textit{feature bounds} that encode the current knowledge about the input, and the action space with \textit{information gathering actions} that refine the feature bounds by performing the same test a DT would do: comparing a feature value to a threshold. The reward function is also augmented with a penalty to take the cost of information gathering action into account. From here on out, when mentioning the interpretability-performance trade-off, we mean the one realized by the IBMDP formulation. A detailed description of IBMDPs is provided in Sec.~\ref{sec:ibmdp}. 
Critically, for and IBMDP policy to yield a tree, it must \begin{enumerate*}\item only depend on the feature bounds part of the state\label{enum:first}, \item not depend on a history of feature bounds\label{enum:second}, \item be deterministic\label{enum:third}.\end{enumerate*} 

To learn such memoryless policy depending on feature bounds in an IBMDP, \citet{IBMDP} modify existing deep RL algorithms such as PPO \cite{Schulman17} and DQN \cite{mnih2015human} to have a policy network that depends only on the observation, and to have what they refer to as an omniscient critic, that depends on the full state of the IBMDP. Leveraging this framework, the main contributions of our paper are to:\\
\textbf{Sec. \ref{sec:AAC}}. We make a connection between the algorithms proposed in \cite{IBMDP} and the literature on asymmetric actor-critics \cite{Baisero}, showing that a similar modification to TRPO \cite{Schulman15} indeed follows the policy gradient theorem in IBMDPs. \\
\textbf{Sec. \ref{sec:xptoy}}. We show that despite this solid foundation neither this modification to TRPO nor the proposed algorithms of \citet{IBMDP} can learn in IBMDPs where the underlying MDP describes a simple supervised learning problem. \\
\textbf{Sec. \ref{sec:understanding}}. After designing an exact version of \citet{IBMDP}'s algorithm, we provide an in-depth ablation study analysing the reasons of the failure. We observe that approximating either $Q^\pi$ or $\pi$ is the main reason for failure.\\
\textbf{Sec. \ref{sec:erpi}}. We show that using entropy regularization policy iteration mitigates approximation errors. And we show if the underlying MDP describes a supervised learning problem, the partially observable IBMDP can be reformulated into a fully observable MDP, alleviating challenges \ref{enum:first}-\ref{enum:third} and making such policy iteration scheme optimal w.r.t to the interpretability-performance trade-off.\\ 
\textbf{Sec. \ref{sec:real-data}}. We perform an experimental comparison between the entropy regularized policy iteration algorithm and the well-known CART algorithm for DT induction. 

\textbf{The take home message of the paper is that optimizing within the discrete space of DTs can be realized efficiently using dynamic programming with convergence guarantees for finding a DT that optimally trades-off between interpretability and performance}. 
 \section{Preliminaries}
 \begin{figure*}[]
\vskip -0.2in
    \hfill
    \subfloat[\tiny{Initialisation of the IBMPD}\label{fig:a}]{%
      \includegraphics[width=0.23\textwidth]{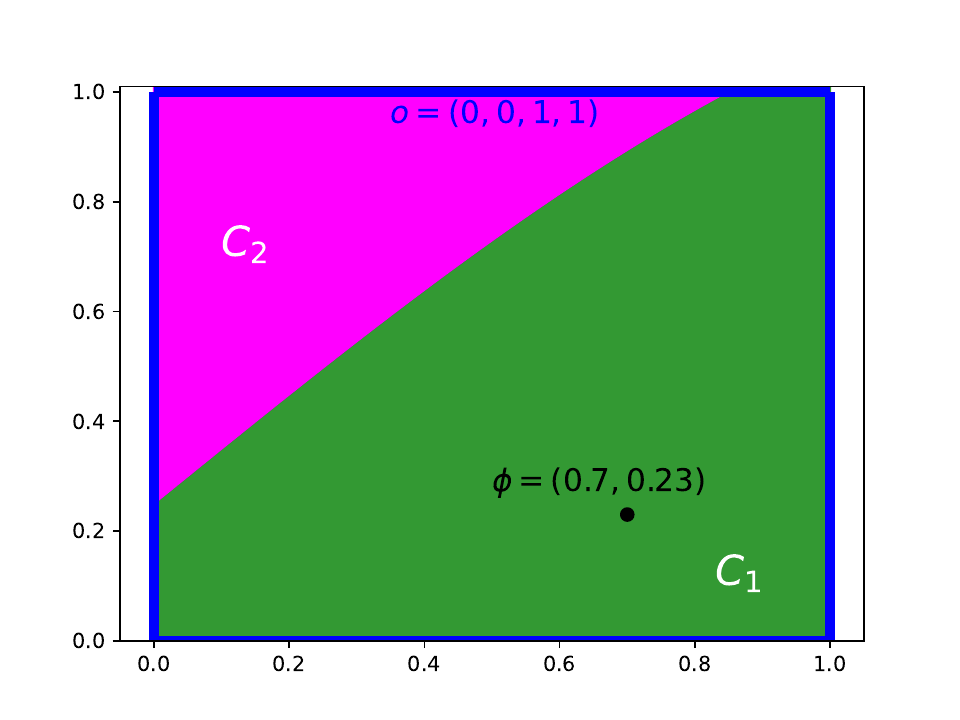}
    }
    \hfill
    \subfloat[\tiny{We take IGA $(\phi_2, 0.5)$ and receive reward $\zeta$}\label{fig:b}]{%
      \includegraphics[width=0.23\textwidth]{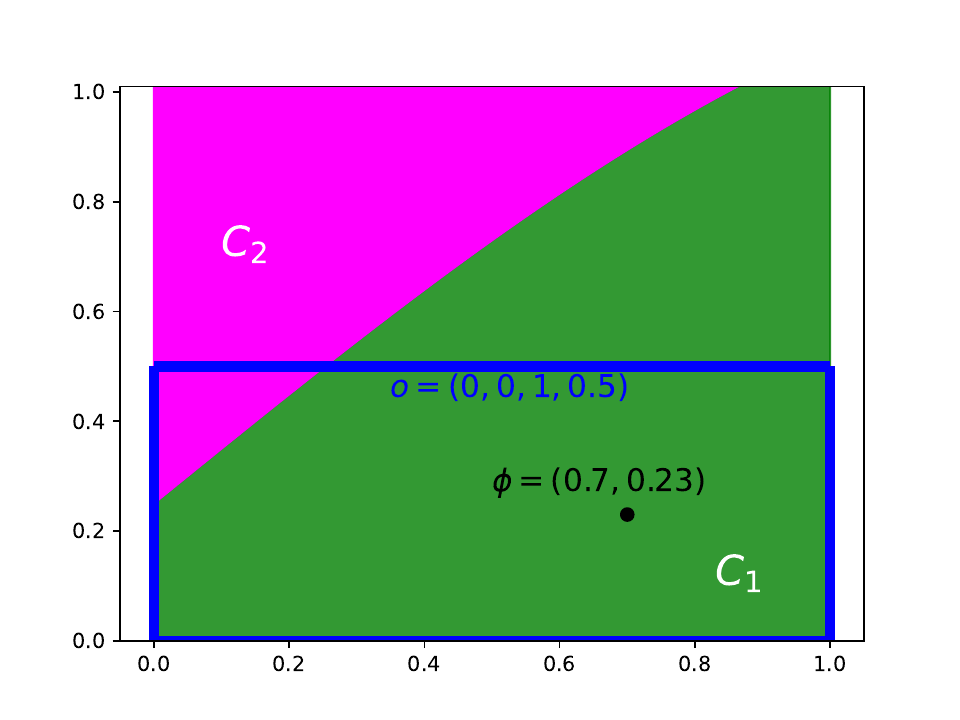}
    }
    \hfill
    \subfloat[\tiny{Taking an IGA in the IBMDP adds a decision node to a DT}\label{fig:c}]{%
      \includegraphics[width=0.23\textwidth]{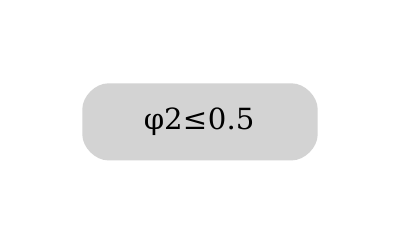}
    }
    \\
    \vskip -0.2in
    \subfloat[\tiny{We take IGA $(\phi_2, 0.25)$ and receive reward $\zeta$}\label{fig:d}]{%
      \includegraphics[width=0.23\textwidth]{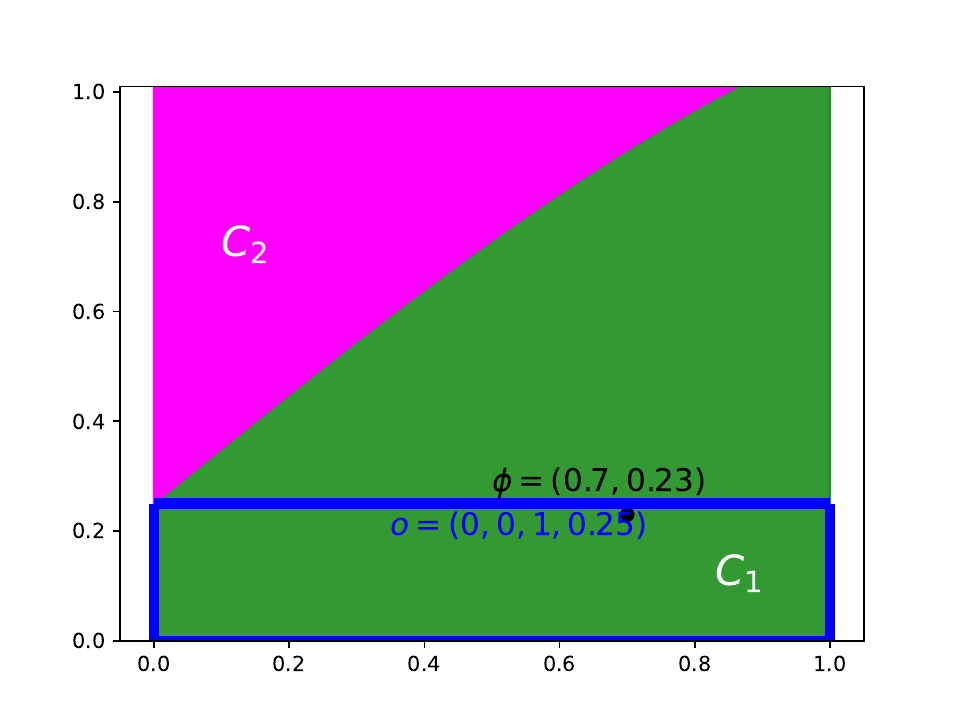}
    }
    \hfill
    \subfloat[\tiny{Taking an IGA in the IBMDP adds a decision node to a DT}\label{fig:e}]{%
      \includegraphics[width=0.23\textwidth]{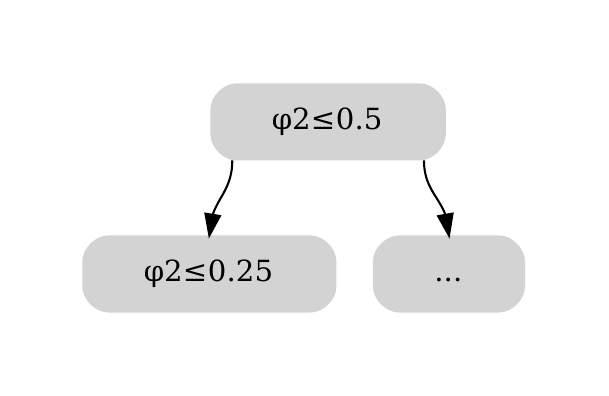}
    }
    \hfill
    \subfloat[\tiny{We take base action $C_1$ and receive reward 1 as $\phi \in C_1$}\label{fig:f}]{
      \includegraphics[width=0.23\textwidth]{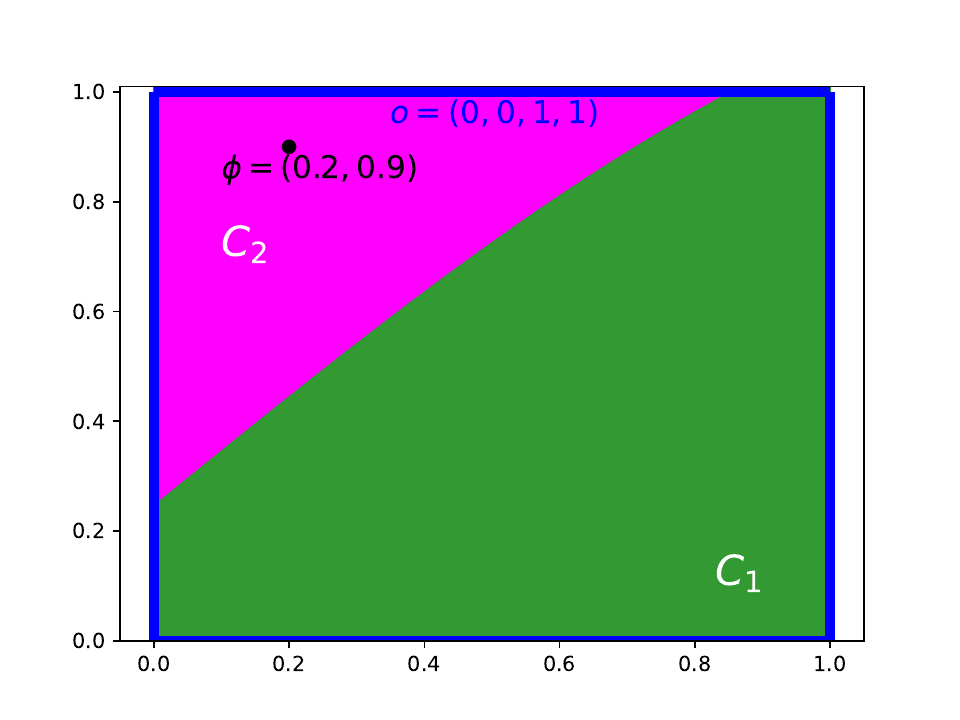}
    }
    \hfill
    \subfloat[\tiny{Taking a base action add a decision node to the DT}\label{fig:g}]{%
      \includegraphics[width=0.23\textwidth]{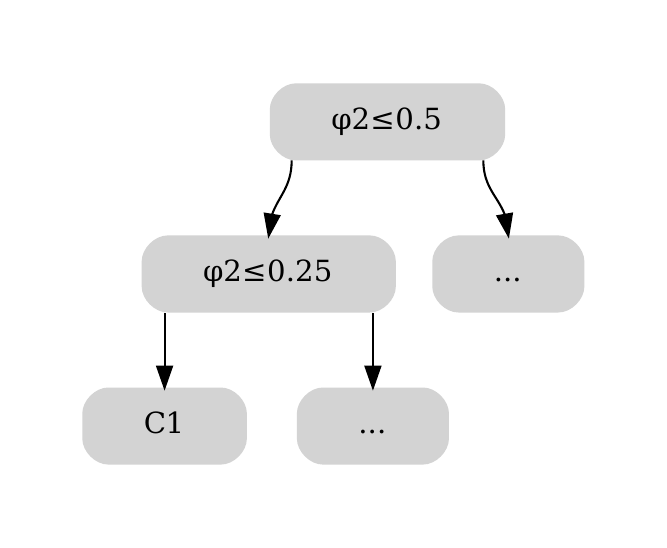}
    }
    \caption{Example IBMDP trajectory: at first in \ref{fig:a} the IBMDP is initialised: the base state $\phi$ is drawn at random from the base MDP and the feature bounds $o$ are set to $(0,0,1,1)$. Then, in \ref{fig:b} an agent takes the IGA $(\phi_2 ,0.5)$; the base part $\phi$ of the state remains unchanged but the observation part is updated to $ o = (0,0,1,0.5)$ because $\phi_2 = 0.23\leq 0.5$. Antother IGA is taken in \ref{fig:d}. Finally, in \ref{fig:f}, an agent takes a base action, so a new base state $\phi$ is drawn from the base transition function and the feature bounds are reset: $o = (0,0,1,1)$.}
    \label{fig:dt-ibmdp1}
  \end{figure*}
In this section, we briefly introduce the notations used throughout the paper, and explain the relation between IBMDPs and the learning of decision trees.

\subsection{Markov decision processes}
We consider an infinite horizon MDP \cite{puterman} defined by the tuple $\langle \SC, \A, R, T, \gamma \rangle$, where $\SC$ is the state space, $\A$ is the discrete action space, $R: \SC \times \A \mapsto [R_{\min}, R_{\max}] \subset \mathbb{R}$ is the bounded reward function, $T$ is the transition function, and $\gamma<1$ is the discount factor. The agent interacts with the environment according to its policy $\pi$. At time $t$, 
the agent takes action ${a}_t \sim \pi(. |{s}_t), \ {a}_t \in {\A}$, after which it observes the reward $r_t$ and the next state ${s}_{t+1}$ with probability $T({s}_t, {a}_t, {s}_{t+1})$. 
Let $Q^\pi(s, a) = \mathbb{E}_{\pi} [\sum_{t\ge{}0} \gamma^t R({s}_t, {a}_t) \mid {s}_0 = s, a_0 = a]$ be the Q-function,  $V^\pi(s)=\mathbb{E}_{\pi}[Q(s,a)]$ be the value function, $A^\pi(s,a) = Q^\pi(s, a) - V^\pi(s)$ be the advantage function, and $J(\pi) = \mathbb{E}[V(s_0)]$ be the policy return for some initial state distribution.
Our goal is to find a policy $\pi$ that maximizes $J$.


\textbf{Partially observable MDPs.} Infinite horizon Partially Observable MDPs (POMDPs) \cite{PDMIA} are tuples $\langle \SC,\Omega, \A, R, T, O, \gamma \rangle$. They are MDPs with an additional observations space $\Omega$ and observation map $O: \SC \rightarrow \Delta \Omega$ that maps states to a distribution over observations. A policy $\pi$ is reactive if it maps an observation to a distribution over actions: $\pi: \Omega \rightarrow \Delta \A$. 

\subsection{Iterative bounding MDPs}\label{sec:ibmdp}

Following \citet{IBMDP}, we introduce the notion of an Iterative Bounding MDP (IBMDP). Let us consider a base MDP $\langle \SC, \A, R, T, \gamma \rangle$. 
 We assume $\SC = [0, 1]^d$. An IBMDP $\langle \SC^{\prime}, \A^{\prime}, R^{\prime}, T^{\prime}, \zeta, p, \gamma\rangle$ is defined on top of it with the following properties.

\textbf{State space} $\SC^{\prime} \subsetneq \SC \times [0, 1]^{2\times d}$. A state $s\in \SC^{\prime}$ has two parts. A base factored state $\phi = (\phi_1, ..., \phi_d)\in \SC$, and a lower and upper bound~$(L_k, U_k)$ for each of the $d$ features---hence requiring $2\times d$ values in $[0,1]$. For each feature $\phi_k$, $(L_k, U_k)$ represents the current knowledge about its value. Initially, $(L_k, U_k) = (0, 1)$ for all $k$, which are iteratively refined by taking Information-Gathering  Actions (IGAs).

\textbf{Action space} $\A^{\prime} = \A \cup \A_I$. An agent in an IBMDP can either take a base action $a \in \A$, or an IGA in $\A_I = \{1,\dots, d\} \times \{\frac{1}{p+1}, ..., \frac{p}{p+1}\}$, with parameter $p \in \mathbbm{N}$.

\textbf{Transition function.} If $a \in \A$, the base state part of an IBMDP state will transition according to the base MDP's transition function $T$, while feature bounds are reset to $(0, 1)$. If $a\in \A_I$, the base state is left unchanged, but the feature bounds are refined. Let $\phi_k$ be the value of the k-$th$ feature of the base state, and $(L_k, U_k) \in [0, 1]^2$ be the current bounds of $\phi_k$. The information gathering action $a = (k, v)$ will compare $\phi_k$ to $v' = v \times (U_k - L_k) + L_k$, and will set the lower bound $U_k$ to $v'$ if $\phi_k > v'$, otherwise $L_k$ is set to $v'$. 

\textbf{Reward function.}
The reward for a base action in $\A$ is defined by the base MDP's reward function $R$. For an IGA in $\A_I$ the reward is a fixed value $\zeta \in (-\inf, R_{\max})$ (the maximum value of the base reward function). We impose $\zeta < R_{\max}$, as otherwise a policy never taking any base action would always be optimal, though this restriction is not enough to prevent this degenerate case for RL algorithms.


\subsection{Learning a decision tree in a IBMDP} \label{sec:learning-dt}
The policy optimization will not occur on the IBMDP but on a partially observable transformation thereof. In this POMDP, the observation space is $\Omega = [0, 1]^{2\times d}$ and the observation map is deterministic $O: \SC \mapsto \Omega$ and simply removes the base state part of an IBMDP state. In the following we will see an IBMDP's state $s\in \SC'$ as a tuple~$(\phi, o)$ with a base state $\phi \in \SC$ and an observation---the feature bounds---$o \in \Omega$. To produce a DT, a policy $\pi$ acting in this POMDP needs to be reactive and deterministic, i.e. $\pi: \Omega \mapsto \A'$. Indeed, in a DT's internal node: \ref{enum:first} the test does not depend on the input (the base state in an IBMDP) and hence, $\pi$ should be a function of $\Omega$. \ref{enum:second} The test does not depend on the identity of the previous node (which is unknown when transitioning back to a root node) and so $\pi$ should be history-less. \ref{enum:third} The same test or base action is taken at every node and hence, $\pi$ should be deterministic. 

Let $\Pi$ denote the space of policies complying with the above restrictions. The learning problem is to find    $\pi^* = \arg\max_{\pi \in \Pi} J(\pi).$
Given $\pi^*$, one can easily extract a DT by simply querying $\pi^*$ on all encountered observations (please see \citet{IBMDP}'s Alg. 1 and Fig.~\ref{fig:dt-ibmdp1}). Let $o_0$ be the observation corresponding to the root of the tree, i.e. where all feature bounds are $(0, 1)$. Assume $\pi^*(o_0) = (k, v) \in \A_I$ is an IGA, then two child nodes $o_1$ and $o_2$ are added to the tree corresponding to the case where either the lower bound or the upper bound of feature $k$ is set to $v$. The extraction continues recursively on $o_1$ and $o_2$ until $\pi^*$ picks a base action on all leaf nodes of the tree. We note that this extraction algorithm will not terminate otherwise. In Sec.\ref{sec:xptoy}, a maximum depth is added to IBMDPs to ensure that a DT can always be extracted from $\pi^*$. A class of RL algorithms that can learn $\pi^*$ is presented in the following section.

\section{Asymmetric actor-critic algorithms} 
\label{sec:AAC}
\subsection{Theoretical grounding}
\begin{figure*}[]
    \vskip -0.2in
    \subfloat[\label{fig:deep}]{%
      \includegraphics[width=0.23\textwidth]{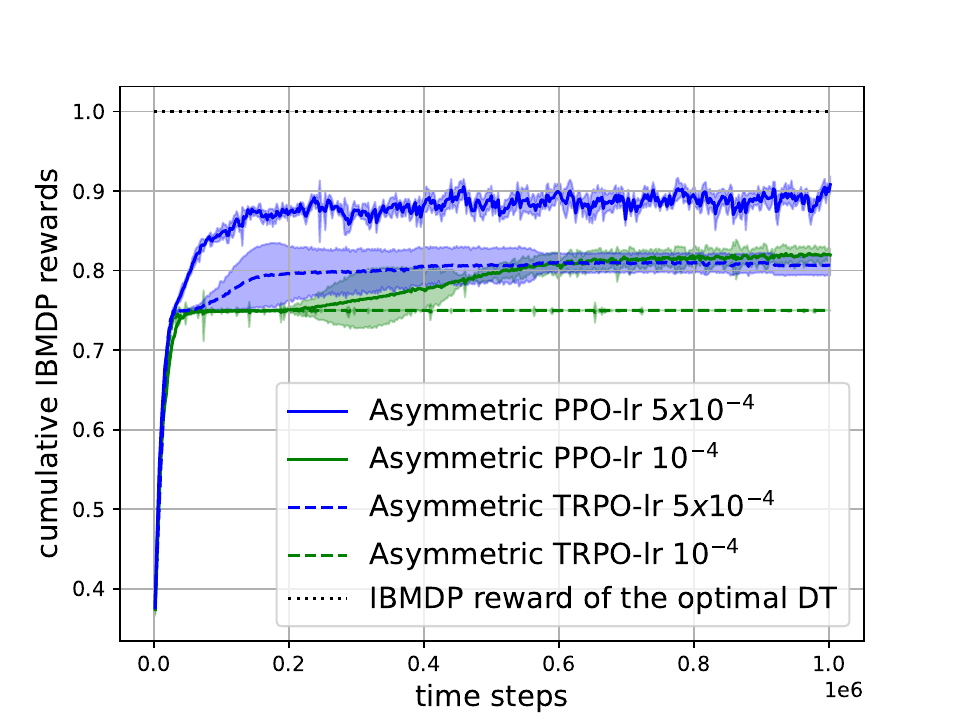}
    }
    \hfill
    \subfloat[ \label{fig:exact-deep}]{%
      \includegraphics[width=0.23\textwidth]{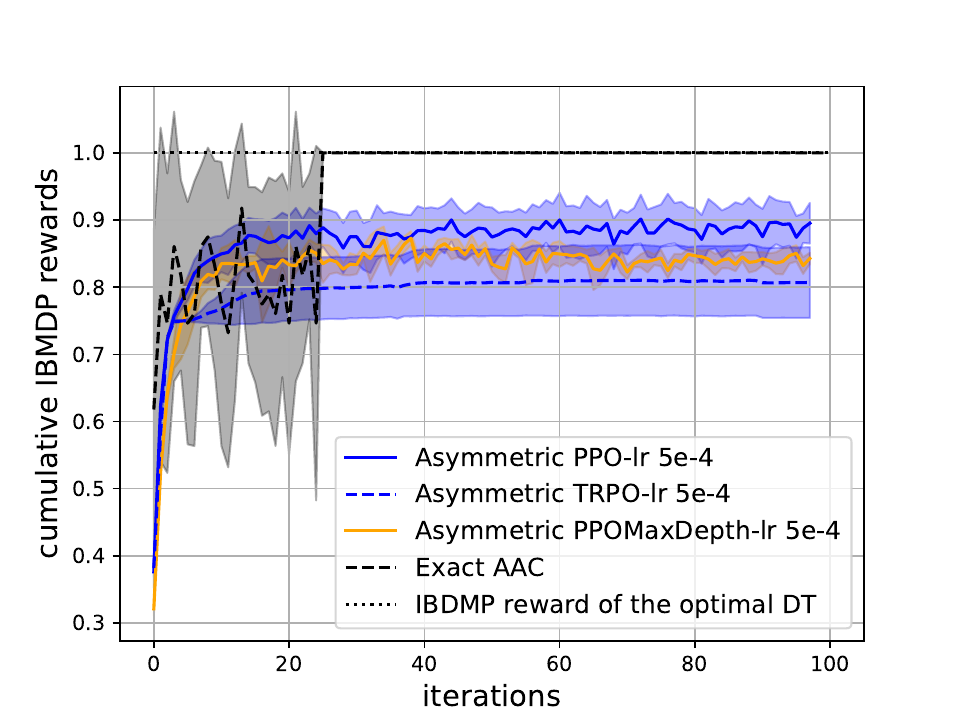}
    }
    \hfill
    \subfloat[\label{fig:ablation}]{
      \includegraphics[width=0.23\textwidth]{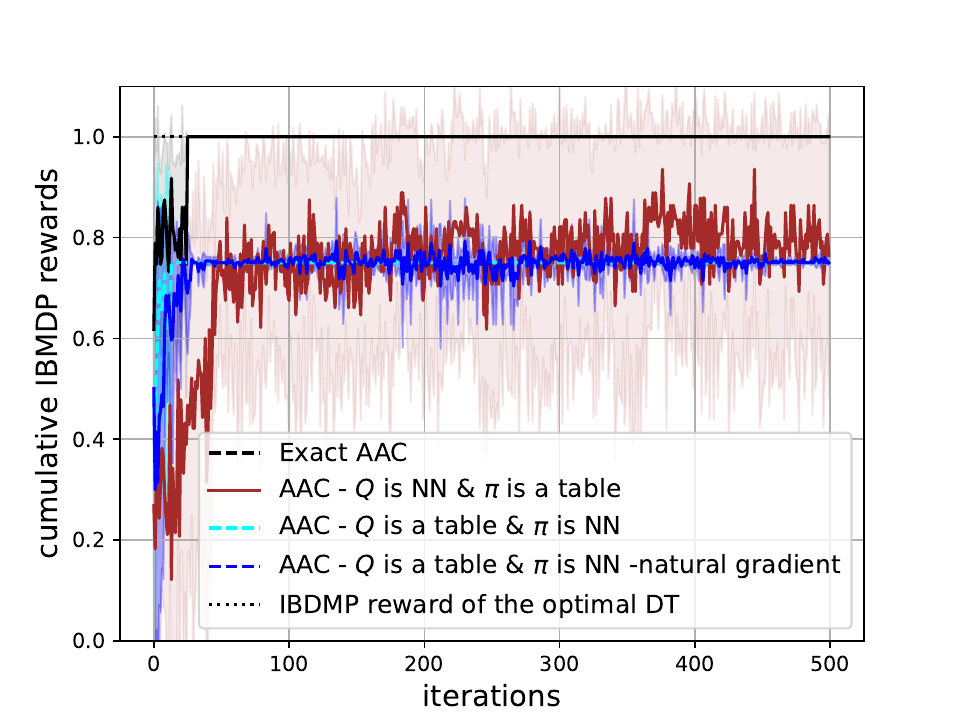}
    }
    \hfill
    \subfloat[\label{fig:qnoisy}]{%
      \includegraphics[width=0.23\textwidth]{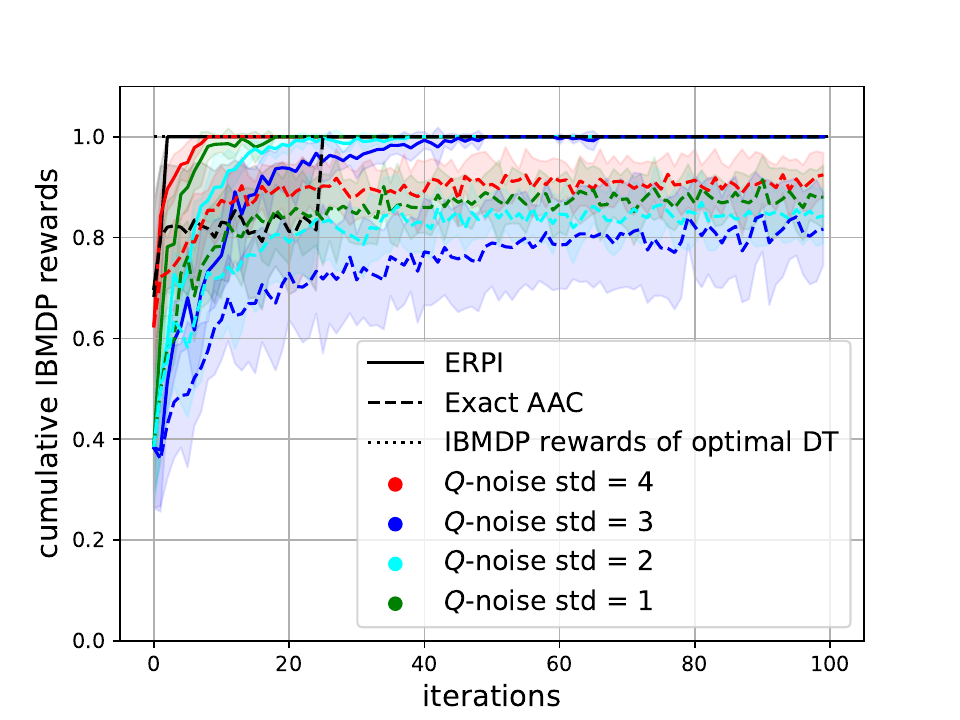}
    }
    \caption{Study of AAC algorithms ability to retrieve DTs for simple supervised classification tasks by learning a reactive policy maximizing the cumulative reward of IBMDPs. See Sec. \ref{appendix:supibmdp} for details about the IBMDPs}
    \label{fig:AAC}
  \end{figure*}
To learn $\pi^*$, \citet{IBMDP} modify PPO, an actor-critic algorithm, to allow the critic to use the full state information $s = (\phi, o)\in \SC'$, while the policy only depends on $o\in \Omega$. A similar modification to DQN~\cite{mnih2015human} was introduced, with a Q-function depending on $s$, and an actor derived from a second Q-network that only depends on $o$. Although not explicitly discussed in \citet{IBMDP}, these actor-critic architectures are very reminiscent of Asymmetric Actor-Critic (AAC) algorithms of the POMDP community \cite{Pinto, Baisero}, where the asymmetry is between the information fed to the actor and the critic. One of the main result to support learning in IBMDPs is the validity of the policy gradient theorem. Let $\pi_\theta : \Omega \mapsto \Delta\A'$ be a parameterized policy such that $\pi_\theta(a|o)$ is differentiable w.r.t. $\theta$ for any~$(o, a)$. Let $Q^{\pi_\theta} : \SC'\times \A'\mapsto \mathbb{R}$ be the Q-function of $\pi_\theta$ in the IBMDP. Since an IBMDP \textit{is} an MDP---which is clear from its definition in Sec.~\ref{sec:ibmdp}---then 
\begin{align}
\label{eq:polgrad}
    \nabla_{\theta} J(\pi_\theta)
    &=\frac{1}{1-\gamma}{\mathbb{E}}\left[\nabla_{\theta} \log \pi_{\theta}\left(a  \mid o \right) Q^{\pi_{\theta}}(s, a)\right],
\end{align}
is the gradient of the return function at parameter $\theta$. The expectation here is on the state-action space with random variable $s = (\phi, o)$ following the $\gamma$-discounted state distribution of $\pi_\theta$, given by $p^{\pi_\theta}(s) = (1-\gamma)\sum_{t\geq 0}\gamma^t P(s_t=s|s_0, \pi_\theta)$, and $a$ sampled from $\pi_\theta(.|o)$. The proof of Eq.~\eqref{eq:polgrad} follows immediately from the policy gradient theorem, since it is not required that $\pi_\theta$ depends on the full state information as long as it is differentiable w.r.t. $\theta$, see e.g. Theorem 1 of~\cite{Mei20}. The modified PPO of \citet{IBMDP} has an AAC structure but its policy update does not follow the policy gradient in Eq.~\eqref{eq:polgrad} because classical PPO itself does not follow the policy gradient. In the following sections, in addition to asymmetric PPO, we will consider an AAC version of TRPO. TRPO updates the policy following the natural gradient computed from Eq.~\eqref{eq:polgrad} which is thus guaranteed to follow an ascent direction for $J$. 

However, this AAC formulation is for stochastic policies while the algorithm must return a deterministic policy, and it is known that in general, a POMDP does not have an optimal policy that is deterministic \cite{PDMIA}. More broadly, there is not a large body of work for learning reactive policies in a POMDP. Earlier works motivated reactive policies by the computational advantages of considering the shortest possible history length of observations, and convergence guarantees for gradient descent methods to a local optimum were provided in \citet{jaakkola1993convergence, jaakkola1994reinforcement}. With the increase in computational power, this motivation is perhaps not as compelling and learning reactive policies remains an open problem \cite{Lazaric}. This begs the question of how feasible it is to learn in the IBMDP framework, and what is the quality of the obtained tree? 

The promise of IBMDPs to transform the discrete optimization problem over the space of DTs into a differentiable one through the use of deep RL is appealing. However, the optimality of the obtained DTs is not discussed theoretically nor investigated experimentally in \citet{IBMDP}. We propose to investigate this point on supervised classification tasks in the following sections.

\subsection{Experimental evaluation on toy IBMDPs}
\label{sec:xptoy}
\textbf{Task definition.}
In this subsection we answer the question: can AAC algorithms such as asymmetric PPO \cite{IBMDP} or asymmetric TRPO \cite{Baisero} find the optimal policy with respect to the IBMDP objective?  To that end, we design toy experiments that are amenable to their analysis thanks to their very small size.
The base tasks are binary supervised classification tasks with 16 different data points and two numerical features in $[0, 1]$. Each data can be perfectly classified using a depth-2 balanced binary tree. We choose $\zeta = 0.5$, $\gamma = 0.99$ and $p=1$ and verified that the reactive policy equivalent to the balanced depth-2 DT is optimal with respect to the IBMDP objective. For more details about the task definition see appendix \ref{appendix:toytask}.

\textbf{Experimental results.} Deep AAC algorithms, use \texttt{stable-baselines3} \cite{stable-baselines3} implementations of TRPO and PPO. The actor network is modified to only take feature bounds as inputs while the critic network uses the full state. We use default hyperparamters and test two learning rates $\{10^{-4}, 5.10^{-4}\}$. We use 5 independent runs for each of the 7 different IBDMPs and normalize returns on each IBMDP so that results can be aggregated. Fig.~\ref{fig:deep} shows that none of the agents were able to consistently find the best DT~(presented on Fig.\@ \ref{fig:dt-ibmdp}) despite the extreme simplicity of the task.

\section{Understanding failure of asymmetric actor-critic algorithms}\label{sec:understanding}
To better understand how theoretically sound AAC algorithms \cite{Baisero} fail to retrieve optimal DTs for simple supervised classification task, we start from an exact version of AAC algorithms where the $Q^{\pi}$ and policy gradient are computed exactly, which is possible in this tabular setup presented next.
\subsection{Deriving an exact AAC algorithm}
We introduce a variant of an IBMDP that enforces a maximum depth of the resulting DTs---and ensures that the DT extraction algorithm always terminates. Let this maximum depth be $M + 1$. $M$ is the maximum number of consecutive time-steps during which a policy can select an IGA. We implement this by forcing the policy to take a base action each time it has performed $M$ consecutive IGAs. Interestingly, if $p+1$ is prime (where $p$ is the parameter controlling splitting thresholds in IBMDPs), the state space already provides such information to the policy:
\begin{proposition}
\label{prop:prime}
For an IBMDP, if $p+1$ is prime then there is a mapping $\Omega \mapsto \mathbb{N}$ that maps any feature bound to the number of consecutive IGAs taken since the last base action.
\end{proposition}
In other words, the number of consecutive IGAs since the last base actions is directly encoded in the feature bounds~(please see appendix~\ref{app:proofs} for proofs of this and all future statements). Thus we can compare on the same IBMDP algorithms that enforce a maximum tree depth to those of \citet{IBMDP} that do not.

Having fixed a maximum tree depth $M + 1$, the number of unique feature bounds, i.e.\@ the cardinality of the observation space $|\Omega|$ becomes finite and is at most $(2pd)^M$. Here $pd = |\A_I|$ is the number of IGAs available at any time~(if available at all) and the factor of $2$ stems from the two possible state transitions following an IGA. Since the state-action space of an IBMDP becomes finite, and its transition and reward functions are known, one can compute the gradient in Eq.~\eqref{eq:polgrad} exactly. This will let us investigate whether the sub-optimal performance of AAC algorithms is due to approximation errors---e.g. introduced by the learned value function---or if it is a limitation of the gradient descent approach in itself. 

Because $\Omega$ is finite, we can additionally implement policy gradient on tabular policies which would eliminate any representation error of the policy. With a slight abuse of notation, we let in this case $\theta(o,a)$ be the logit of observation-action pair $(o, a)$, i.e. $\pi(a|o)\propto \exp(\theta(o,a))$. By a straightforward application of the chain rule on Lemma~C.1 of \cite{Agarwal19} we obtain:
\begin{proposition} 
\label{prop:grad}
Let $\theta \in \mathbb{R}^{\Omega\times\A'}$ be the logits of a tabular reactive policy of the IBMDP, then:
    \begin{align}
\label{eq:polgradtab}
    \frac{\partial J(\pi_\theta)}{\partial \theta(o, a)}
    &={\sum_{s\in \SC'}} \mathbbm{1}_{O(s) = o}\frac{p^{\pi_\theta}(s)}{1-\gamma}\pi_\theta(a|o) A^{\pi_{\theta}}(s, a).
\end{align}
\end{proposition}
Here $\mathbbm{1}_{O(s) = o} = 1$ if the feature bound part of $s$ is $o$, 0 otherwise.
\subsection{Ablation study}\label{sec:abla}
Starting from the exact asymmetric actor-critic algorithm defined above, we perform an ablation study to get to an AAC algorithm similar to \citet{IBMDP, Baisero}. Algorithms are tested on the same IBMDPs as in Sec. \ref{sec:xptoy} presented in appendix \ref{appendix:toytask}. The main features ablated from the exact AAC algorithm are \begin{enumerate*}
    \item Using an approximated $\hat{Q}^\pi$-function instead of a $Q^\pi$-table updated exactly. In that case, $\hat{Q}^\pi$ is a neural network similar to the one in asymmetric PPO and asymmetric TRPO. 
    \item Using neural network for the policy $\pi$ instead of a table. In that case, the policy network is similar to the one in asymmetric PPO and asymmetric TRPO. 
\end{enumerate*} The results of the ablation are presented on Fig.\@ \ref{fig:ablation}. The exact AAC algorithm consistently finds the optimal policy. This is important as it means that the AAC formulation is not the reason for the poor formulation, at least when implemented exactly. In practice however, we find that both the approximation errors of the neural Q-function and the policy representation error hinder performance. Indeed, when the policy is encoded by a neural network in place of a table, the aggregated cumulative IBMDP reward converges to sub-optimal values after just a few iterations. When the $Q^\pi$ function is encoded by a neural network, some instances of the associated AAC algorithm converged to the optimal policy---reflected by the higher standard deviations on Fig. \ref{fig:ablation}, but many did not.

\section{Improving asymmetric actor-critic algorithms with entropy regularization}
\label{sec:erpi}
A first conclusion of the ablation study is that policy networks are prone to representation errors which are perhaps not helped by the discrete nature of $\Omega$, that might not lend itself to easy generalization. However, by limiting the maximum allowed tree depth, we can resort to tabular policies as there is no dependence of the latter to the continuous part of the IBMDP state. 

Secondly, even if AAC consistently finds the optimal reactive policy in its exact form, the convergence of policy gradient of softmax policies is known to be very slow when the probability of the optimal action gets too low \cite{Mei20}. This is due to the weighting in Eq.~\ref{eq:polgradtab} by $\pi_\theta(a|s)$. What is specific to our case is that $\zeta$ is typically set such that IGAs are bad unless there is a good subsequent sequence of actions that lead to a substantial increase in classification accuracy. Since at the beginning, this sequence is unknown, we have noticed that often at the root node the action probability of a base action is increased greatly at the detriment of IGAs. While exact AAC is able to recover from this after a period of reward stagnation, perhaps the added error in the Q estimation makes the 'momentum' cause by the low IGA probabilities too hard to overcome; and in fact, we have observed that a common failure of asymmetric PPO is to converge to zero depth trees that take no IGA at all.

To increase robustness to noisy Q estimates, a known remedy is entropy regularization that is known to average errors in Q instead of summing them \cite{Geist}. To obtain an asymmetric entropy regularized algorithm, the modification to Eq.~\eqref{eq:polgradtab} is quite minimal: it consists in mainly removing the problematic weighting $p^{\pi_\theta}(s)\pi_\theta(a|o)$. A theoretically justified weighing scheme to aggregate the different $A^{\pi_{\theta}}(s, a)$ into a function of $o$ and $a$ is to define (by a slight abuse of notation) $A^{\pi_{\theta}}(o, a) = \sum_{s\in \SC'}p^{\pi_{\theta}}(s|o)A^{\pi_{\theta}}(s, a)$, where $p^{\pi_{\theta}}(s|o) \propto p^{\pi_{\theta}}(s)$ if $O(s) = o$ and 0 otherwise. We also define the distribution $p^{\pi_{\theta}}(o) = \sum_{s\in \SC'}\mathbbm{1}_{O(s) = o}p^{\pi_\theta}(s)$. An interesting property of $p^{\pi_{\theta}}(s|o)$ is that it is in fact independent of $\pi_\theta$. This is only valid if the base task is a supervised learning one where the base state distribution is independent of the base action. In this case, while the distribution of visited feature bounds $p^{\pi_{\theta}}(o)$ changes with $\pi_{\theta}$, the distribution of visited states given a feature bound $o$ only depends on $o$ and the initial data distribution of base states. From now on, we will write $p(s|o)$ dropping the policy dependence. This property allows us to express the performance difference lemma \cite{Kakade02} for reactive policies as a function of the above defined quantities $A^{\pi_{\theta}}(o, a)$.

\begin{lemma}
\label{lemma:pdl}
Let $\pi$ and $\pi'$ be two reactive policies of an IBMDP, i.e.\@ mappings of the form $\Omega \mapsto \Delta \A'$, and define $A^\pi(o, a) = \sum_{s\in\SC'}p(s|o)A^\pi(s, a)$ then: \newline $J(\pi') - J(\pi) = \frac{1}{1-\gamma}\sum_{o\in \Omega}\sum_{a\in\A'}p^{\pi'}(o)\pi'(a|o)A^\pi(o, a)$.
\end{lemma}

When replacing the weighting of $A^{\pi_{\theta}}(s, a)$ by $p(s|o)$, in the gradient $\frac{\partial J(\pi_\theta)}{\partial \theta(o, a)}$, the ``gradient'' ascent of the logits becomes:
\begin{align}
    \label{eq:rpiup}
    \theta_{k+1}(o,a) = \theta_k(o,a) + \alpha A^{\pi_{\theta_k}}(o,a),
\end{align}
for all $(o, a)$. Let Alg.\@ 1 an algorithm that learns at every iteration $k$, $A^{\pi_{\theta_k}}$ and updates the policy according to Eq.~\eqref{eq:rpiup}. Alg.\@ 1 is indeed a form of Entropy Regularized Policy Iteration (ERPI), where logits are sums of all past advantage functions~(or equivalently Q functions) \cite{evendar, Geist, Politex}. Let $\pi^*$ be the best deterministic reactive policy as defined in Sec.~\ref{sec:learning-dt}. Combining Lemma~\ref{lemma:pdl} with the usual analysis of ERPI algorithms \cite{evendar, Geist, Politex} one can show:
\begin{theorem}
\label{thm:rpiconv}
For an initial reactive policy $\pi_{\theta_1}$ uniform over $\A'$ for all $o\in\Omega$, after $K$ iterations of Alg. 1 we have
\begin{align*}
    J(\pi^*) - \max_{k\in\{1,\dots, K\}}J(\pi_k) \leq \frac{\sqrt{2\log |\A'|}  ( R_{\max} -R_{\min})}{2\sqrt{( 1-\gamma )^{3} K}}
\end{align*}
\end{theorem}
Theorem~\ref{thm:rpiconv} shows that there exists a tractable algorithm that finds a (stochastic) reactive policy that performs arbitrarily close to $\pi^*$. This is only true for IBMDPs that extend MDPs defining a supervised learning task. For the more general case as described in \cite{IBMDP}, we stress out that this remains an open problem. Interestingly, the convergence rate depends on the number of IGAs (and hence the number of features of the data) but not on $|\Omega|$ and the maximal tree depth. 

The modification of ERPI compared to the gradient update of Eq.~\ref{eq:polgradtab} is minimal, yet the resulting algorithm is much more robust to noisy Q-functions as shown in Fig.~\ref{fig:qnoisy}. This robustness to erroneous Q-functions was also shown from a more theoretical point of view in \cite{Geist}. In general, implementing ERPI is difficult in practice as it requires keeping track of all previous Q-functions \citet{Politex}, which are typically Q-networks in deep RL and which can be too costly. However, as previously discussed, learning a reactive policy introduces difficulties but has the advantage of depending only on the finite set $\Omega$ and not on the continuous base state, which makes ERPI possible in our case as we only need to store $A^{\pi}(o,a)$.

Beyond ERPI, the fact that $p(s|o)$ is independent of the policy yields a more general result. Indeed, one can transform an IBMDP into an MDP, which we call the Observation-IBMDP $\langle \Omega, \A^{\prime}, R'', T'', \gamma\rangle$ defined over the feature bounds space, with $R''(o, a) = \sum_{s\in\SC'}p(s|o)R'(s,a)$ and $T''(o_t, a_t, o_{t+1}) = \sum_{s, s' \in \SC'}\mathbbm{1}_{O(s') = o_{t+1}}p(s|o_t)T'(s, a_t, s')$. The main interest of the Observation-IBMDP is that
\begin{theorem}
\label{theo:obs}
    Any optimal policy of the Observation-IBMDP has the same policy return as $J(\pi^*)$ in the IBMDP.
\end{theorem}
Thus one can use any algorithm to find a deterministic policy in the Observation-IBMDP and use this policy to extract the DT optimizing the interpretability-performance trade-off. In the next section we provide proof of concepts of our DT learning framework using ERPI on toy datasets. We leave it to future work to investigate how to scale to larger datasets using e.g. deep RL, Monte Carlo Tree Search~\cite{MCTS} or a combination of both as in AlphaGo~\cite{alphago}.

To conclude this section, we provide a bound on the performance of the deterministic policy extracted from ERPI: 
\begin{proposition}
\label{prop:deterministic}
Let $\pi = \arg\max_{\pi\in\{\pi_1,\dots, \pi_K\}}J(\pi_k)$ where $\{\pi_1,\dots, \pi_K\}$ is the set of policies generated after $K$ iterations of Alg.~1 (ERPI). The deterministic policy $\pi_D(o) \in \arg\max_{a\in\A'} A^{\pi}(o, a)$ satisfies $J(\pi^*) - J(\pi_D)\leq J(\pi^*) - J(\pi)$.
\end{proposition}

\section{Learning decision trees}\label{sec:real-data}
\begin{figure*}[]
    \vskip -0.2in
    \subfloat[DT obtained by ERPI when $\zeta = -0.6$\label{fig:wine-dt1-erpi}]{
      \includegraphics[width=0.23\textwidth]{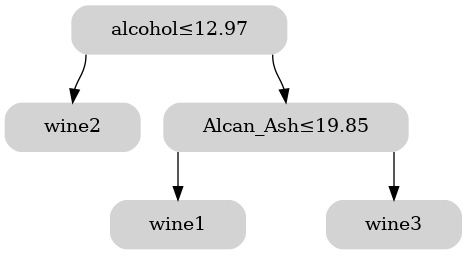}
    }
    \hfill
    \subfloat[DT obtained by ERPI when $\zeta = 0.8$\label{fig:wine-dt2-erpi}]{%
      \includegraphics[width=0.23\textwidth]{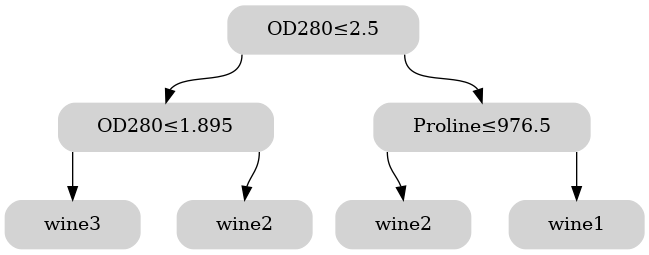}
    }
    \hfill
    \subfloat[DT obtained by CART when $M = 1$\label{fig:wine-dt1-cart}]{
      \includegraphics[width=0.23\textwidth]{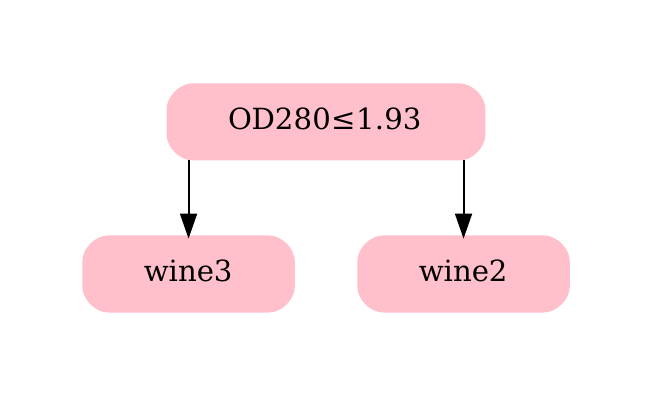}
    }
    \hfill
    \subfloat[DT obtained by CART when $M = 2$\label{fig:wine-dt2-cart}]{%
      \includegraphics[width=0.23\textwidth]{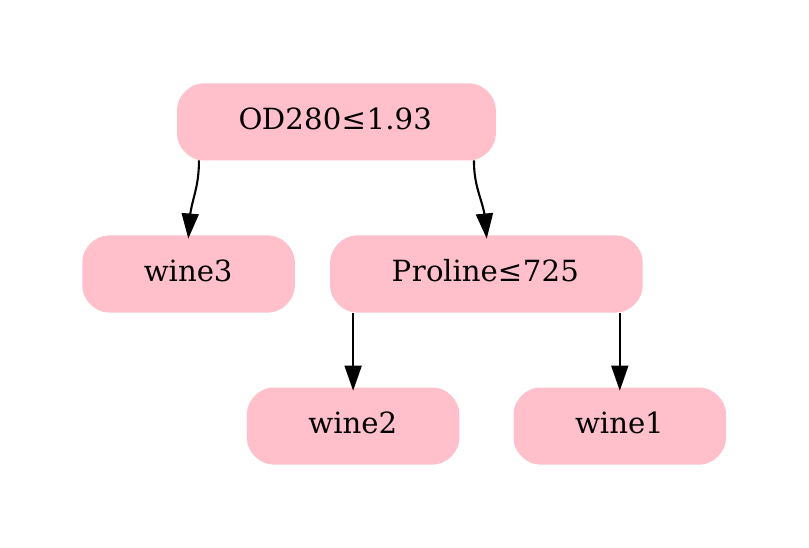}
    }
    \caption{DTs obtained by ERPI and CART on the \texttt{wine} dataset.}
    \label{fig:interesting-trees}
  \end{figure*}
\begin{figure}[]
\vskip -0.1in
    \centering
    \subfloat[Trade-off on \texttt{wine} dataset\label{fig:perf-iris}]{%
      \includegraphics[width=0.45\columnwidth]{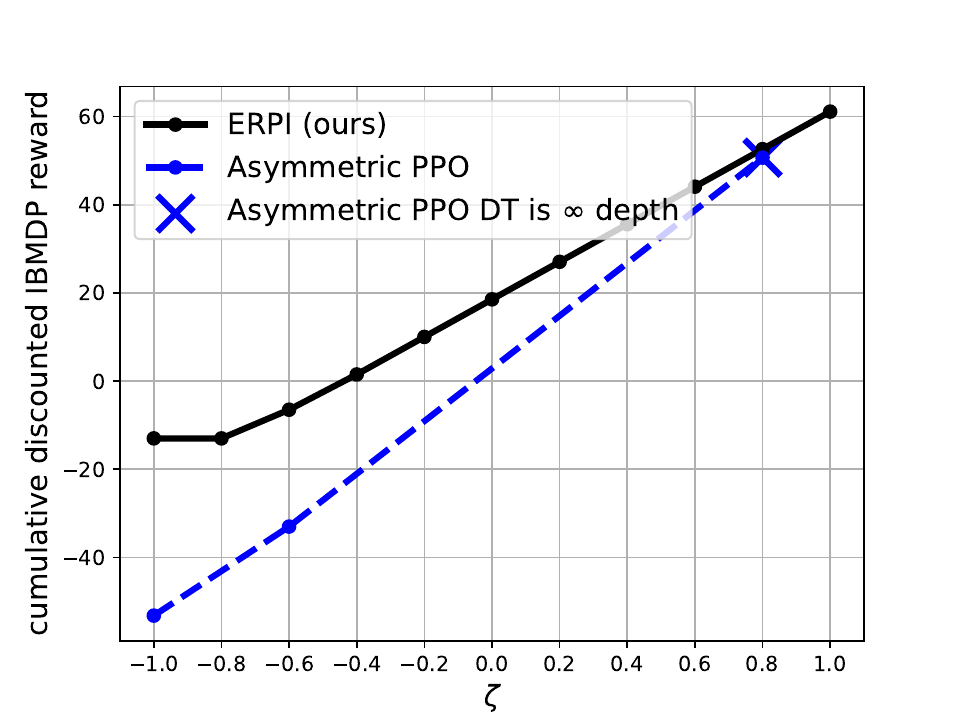}
    }
    \subfloat[Trade-off on \texttt{iris} dataset\label{fig:perf-wine}]{
      \includegraphics[width=0.45\columnwidth]{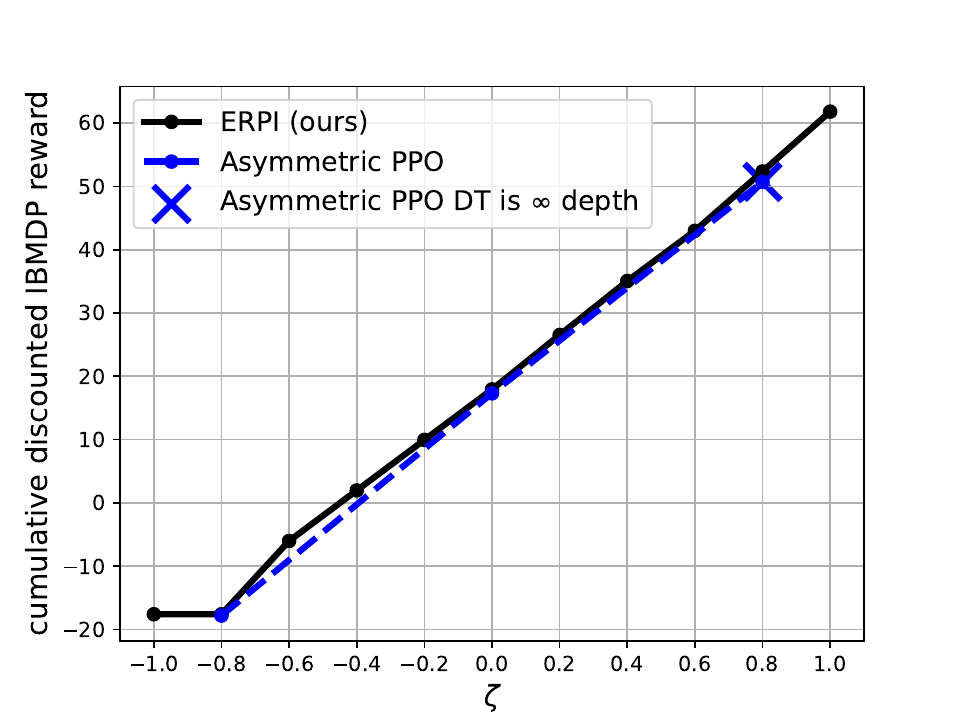}
    }
\caption{Interpretability-performance trade-offs of ERPI and asymmetric PPO on real world datasets as a function of $\zeta$}
    
    \label{fig:perf-inter-trade-off}
  \vskip -0.2in
  \end{figure}
In this section we compare ERPI (Sec.\@ \ref{sec:erpi}) and asymmetric PPO (Sec.\@ \ref{sec:AAC}) with CART, a well-known decision tree learning algorithm. 
We consider two small scale datasets, \texttt{iris} and \texttt{wine}. \texttt{Iris} is made of 150 examples, each having 4 features. \texttt{Wine} is made of 178 examples each having 13 features. In both datasets, there exists 3 classes.

\textbf{ERPI:} we apply ERPI on IBMDPs for all $\zeta \in \{-1, -0.8, ..., 0.8, 1\}$. We set $p=1$ so that the splitting threshold for feature $k$ in the learned DTs is $\frac{U_k - L_k}{2} + L_k$. 

\textbf{Asymmetric PPO:} we apply asymmetric PPO \cite{IBMDP} on IBMDPs of the real world classification taks for all $\zeta \in \{-0.8, 0, 0.8\}$ for \texttt{iris} and for all $\zeta \in \{-1, -0.6, 0.8\}$ for \texttt{wine}. These values for $\zeta$ are the ones that give DTs different from one an other using ERPI. The other parameters of the IBMDPs are the same as for ERPI.

\textbf{CART:} 
CART aims at minimizing the class prediction error rate. 
We use \texttt{scikit-learn} \cite{scikit-learn} implementation of CART. We fix the maximum depth parameter: once the tree has reached the maximum depth, no more split is made and the node is a leaf. In particular, we set the maximum depth to 1, 2, 3 and 4 for \texttt{iris} and to 1 and 2 for \texttt{wine}. Other hyperparameters are the default ones. It is important to note that CART chooses splitting thresholds among features values in the dataset. That is a major difference with ERPI and asymmetric PPO applied on IBMDPs for which splitting thresholds are chosen in the usually smaller action space of the IBMDPs $\A_I$. 

\textbf{Interpretability-Performance trade-off:} on figure \ref{fig:perf-inter-trade-off}, we observe that ERPI's cumulative discounted IBMDP reward is always above the one of asymmetric PPO on both datasets. When both curves are close to each other for $\zeta = 0.8$, asymmetric PPO usually learns an infinite depth DT. We can conclude that in practice, ERPI is better than asymmetric PPO at optimizing the interpretability-performance trade-off which supports the theory of Sec.~\ref{sec:understanding} and  \ref{sec:erpi}. The study of the interpretability-performance trade-off for CART can only be studied qualitatively as CART does not optimize the cumulative discounted IBMDP reward. In general, since we have chosen a small value of $p$ for the IBMDP, CART typically chooses finer splits leading to improved accuracy compared to trees found by ERPI. 
Apart from this difference in the threshold value, it is interesting to see that for \texttt{iris},  the different trees obtained by ERPI when varying $\zeta$ share the same structure and simply split additional nodes as $\zeta$ increases. This suggests that a greedy algorithm such as CART would have done as good a job at optimizing the performance-interpretability trade-off as RL algorithms. On the \texttt{wine} dataset---which has more than 3 times the number of features of \texttt{iris}---the trees found by ERPI for values $\zeta = -0.6$ and $\zeta = 0.8$ chose completely different features. Whereas CART splits the same feature when $M = 1$ and $M=2$. This clearly shows the advantage of doing a full optimization over the set of all possible DTs to find the best interpretability-performance trade-off. In App.~\ref{appendix:trees}, DTs learned by ERPI, asymmetric PPO and CART are shown with their accuracy on each dataset.



\section{Related work}
\textbf{Explainable Reinforcement Learning.}
    Explainable RL refers to explanations about an RL agent's decision making process. There are different levels of explanations ~\cite{Milani}. Feature Importance (FI) explanations provide an action-level look at the agent’s behavior: for each action, one can query for the immediate context that was critical for making that decision. For example, ~\cite{landajuela} use deep RL to directly search the space of few-terms algebraic expressions policies. They successfully learn high-performing algebraic policies that are way less complex than neural network policies on classical control benchmarks (for the CartPole environment they find $\pi(s) = 10s_3 + s_4$). Such FI method was able to highlight the important state features for the control of the CartPole (here, features 3 and 4). 
    
    \textbf{Interpretability in Reinforcement Learning.}
       In the supervised learning community, one can achieve interpretability by either i) learning a black-box model and making it interpretable post-hoc by sensitivity analysis or by (locally) mimicking the black-box with an interpretable model~\cite{Strumbelj10, Ribeiro16, Craven96} or ii) by learning an interpretable model from the get-go such as linear models, decision trees or attention-based networks~\cite{Ustun15, Letham15, Vaswani17}. In RL, a similar categorization exists, where prior work either considered explaining learned policies~\cite{Madumal19, Hayes17}, learning an interpretable policy by imitation learning of a neural network policy~\cite{Verma18, Viper} or learning an interpretable policy from scratch using for example linear policies~\cite{Nie19}, decision trees \cite{DDT, IBMDP} or mixture of experts \cite{akrour21}.

    \textbf{Decision Tree with Reinforcement Learning.}
    Specifically regarding DTs, current work mainly considered trees where each node tests whether the value of a certain numerical feature is smaller than a certain threshold. Related to this setting, there has been prior research on learning such DTs with RL, e.g. using the Fitted Q Iteration (FQI) algorithm~\cite{Ernst05}. However, this work did not consider interpretability and the trees produced by FQI are rather large even on simple problems \cite{Viper}. To learn compact DTs one can either directly use a tree as a policy function approximator or use a neural network as the policy function approximator in a transformed problem. In the first case, one can cite DDT \cite{DDT}, which uses a differentiable decision tree (DDT) as policy function approximator, having a fixed structure (including the choice of state features per node) and uses policy gradient to optimize the splitting value in every node. On the other case, one can transform the base MDP for which we want to learn the DTP following \cite{IBMDP}, and use RL algorithms inspired from the POMDP community to learn a neural policy in this setting.
\section{Discussion}\label{sec:disc}
\textbf{Scaling.}
One of the main contribution of the paper is to show that even though learning reactive policies in an IBMDP is in general hard, the specific setting of learning a DT for a classification task is perfectly feasible by RL. One of the main limitation of our current work is that of scaling, especially with the number of features and the granularity of the splitting, since the size of the state space of the resulting RL problems grows as $(2pd)^M$. However, algorithms such as MCTS \cite{MCTS} are shown to scale to problems with very high branching factors, and are thus an interesting future step to scale our work. One interesting research question is whether the combination of MCTS with deep RL, such as in AlphaGo \cite{alphago}, can be beneficial. While for smaller tasks we have seen that the discrete nature of $\Omega$ was limiting the benefits of generalization, the behavior of neural networks on a much larger $\Omega$ remains to be studied. In all likelihood, a library for optimizing the interpretability-performance trade-off using RL should choose among the spectrum from tabular to deep RL methods depending on the characteristics of the dataset since tabular, fast methods, can still have benefits in allowing a quick exploration of several values of $\zeta$.

\textbf{Beyond DTs.} Scaling to larger IGA spaces would allow for finer splittings at decision nodes, but could also ease the generalization of IBMDPs to more general classes of interpretable models. At the very least one could consider test nodes that also test if a feature value is within closed intervals $[v_1, v_2]$ or for example could consider combinations of more than one feature. As long as one can write the associated transition function, it is very likely that our results would extend to such settings too. 

\textbf{Beyond supervised learning.}
Finally, we stress out that provably convergent algorithms for finding an optimal deterministic reactive IBMDP policy when the base task is a sequential decision making problem remains an open question. Perhaps one intermediate step is to study the learning of DT policies by imitation learning as in \cite{Viper}, which can be reduced to a sequence of supervised learning problems.

\section{Conclusion}
In this work we analysed and evaluated the recently proposed IBMDP framework \cite{IBMDP} that tackles the interesting problem of learning compact decision trees with RL. We cast the problem in the AAC framework and showed experimentally that AAC algorithms 
fail to retrieve the optimal DT for IBMDPs of simple supervised classification base tasks. One of the main contribution of the paper is to show that this problem can be reformulated into a fully observable MDP and that learning the optimal DT is tractable. This opens up a large avenue for future research to go beyond greedy search algorithms and explore the full space of decision trees and similar models of discrete nature. 

\bibliography{example_paper}
\bibliographystyle{icml2023}

\newpage
\appendix
\onecolumn
\section{Proofs}
\label{app:proofs}
\subsection{Proposition~\ref{prop:prime}: tree depth information in feature bounds}
\label{app:prime}
We want to show that in an IBMDP, if $p+1$ is prime, then the number of information-gathering actions performed since the last base action is encoded in the feature bounds part of the state. Without loss of generality we only study the case of a single feature, showing that $l = U - L$, the difference between the upper and lower bound of the feature fully determines the number of information-gathering actions taken since the last base action. The extension to multiple features is trivial by addition of each feature's inferred number of information-gathering actions. 

Let $n$ be the number of information-gathering actions taken since the last base action. Let $l_n$ be the difference between upper and lower bounds of the feature after these $n$ information-gathering actions. Clearly $n = 0 \Leftrightarrow l_n = 1$. When $n > 0$, $l_n$ is given by $l_n = \overset{n}{\underset{k=1}{\prod}}\frac{h_k}{p+1}$, where $h_k\in \{1,...,p\}$ for each $k$. 
We want to show that if $p+1 \text{ is prime} \implies \nexists (i,j): i < j \text{ and } l_i = l_j$.
\begin{align*}
    \text{Suppose } &\exists (i,j): 0<i<j \text{ and } l_i=l_j,\\
    & \implies \frac{\overset{i}{\underset{k=1}{\prod}}h_k}{(p+1)^{i}} = \frac{\overset{j}{\underset{k=1}{\prod}} h'_k}{(p+1)^{j}},\\
    & \implies \overset{i}{\underset{k=1}{\prod}}h_k = (p+1)^{j - i} \overset{j}{\underset{k=1}{\prod}} h'_k,  \\
    & \implies \overset{i}{\underset{k=1}{\prod}}h_k = (p+1) (p+1)^{j - i - 1} \overset{j}{\underset{k=1}{\prod}} h'_k,\\
\end{align*}
But that is impossible since the left-hand side is the product of non-zero natural numbers $< p + 1$ and the right-hand side is the product of non-zero natural numbers containing the prime number $p+1$. $\square$

\subsection{Proposition~\ref{prop:grad}: gradient of soft-max reactive policy}
\label{app:gradtab} 
For an MDP $\langle \SC, \A, R, T, \gamma \rangle$ with finite state-action spaces, Lemma~C.1 of \cite{Agarwal19} showed that for tabular soft-max policies
\begin{align}
    \label{eq:tabmdp}
    \frac{\partial J(\pi_\theta)}{\partial \theta(s, a)}
    &=\frac{1}{1-\gamma}p^{\pi_\theta}(s)\pi_\theta(a|s) A^{\pi_{\theta}}(s, a),
\end{align}
where $\theta(s, a)$ is the logit parameter of the policy. Now extending this MDP into an IBMDP $\langle \SC^{\prime}, \A^{\prime}, R^{\prime}, T^{\prime}, \zeta, p, \gamma\rangle$ with a fixed maximum depth as defined in Sec.~\ref{sec:understanding}, the state space remains finite. Let $s = (\phi, o) \in \SC'$ be a state of the IBMDP, with $o \in \Omega$ where $\Omega$ is the finite set of reachable feature bounds. We are interested in tabular policies parameterized by logits $\theta' \in  \mathbb{R}^{\Omega \times \A'}$ such that $\pi_{\theta'}(a | s) \propto \exp(\theta'(o, a))$. That is, the main difference with the setting of Eq.~\eqref{eq:tabmdp} is that a given logit $\theta'(o, a)$ is shared between several states and provides the unormalized log-probability of taking action $a$ in all states $s \in \SC'$ such that their feature bounds is $o$, i.e. such that $O(s) = o$. Informally, we can decompose this map $\theta' \mapsto J(\pi_{\theta'})$ going from logits in $\mathbb{R}^{\Omega \times \A'}$ to a policy return into the composition $\theta' \mapsto \theta \mapsto J(\pi_\theta)$, where the first map maps logits in $\mathbb{R}^{\Omega \times \A'}$ into logits in $\mathbb{R}^{\SC' \times \A'}$ according to $\theta(s, a) = \theta'(O(s), a)$. By the chain rule, we have $\nabla_{\theta'}J(\pi_{\theta'}) = H(\theta, \theta')^T\nabla_{\theta}J(\pi_\theta)$ where $H$ is the Jacobian of the map $\theta'\mapsto \theta$. This Jacobian will have a value of 1 at row $(s, a)$ and column $(O(s), a)$ for all $s$ and $a$ and is 0 otherwise. Thus the product simply becomes
\begin{align}
\label{eq:polgradtab_int}
    \frac{\partial J(\pi_\theta)}{\partial \theta'(o, a)}
    &={\sum_{s\in \SC'}} \mathbbm{1}_{O(s)=o}\frac{\partial J(\pi_\theta)}{\partial \theta(s, a)}\bigg|_{\theta(s, a) = \theta'(o, a)}
\end{align}

Combining Eq.~\eqref{eq:tabmdp} and Eq.~\eqref{eq:polgradtab_int} completes the proof of Eq.~\eqref{eq:polgradtab}.

\subsection{Lemma~\ref{lemma:pdl}: performance difference as a function of $A^\pi(o, a)$}
Let $\pi$ and $\pi'$ be two reactive policies of an IBMDP, i.e. maps of the form $\Omega \mapsto \Delta \A'$. The performance difference lemma \cite{Kakade02} in this case still holds since the feature bounds are part of the state of an IBMDP, and we have
\begin{align}
J\left( \pi'\right) -J ( \pi) & =\frac{1}{1-\gamma }\mathbb{E}\left[ A^\pi(s, a)\middle| s\sim p^{\pi'} ,a\sim \pi'(.|O(s))\right]\\
 & =\frac{1}{1-\gamma } \ \sum _{s\in S'} p^{\pi'}( s) \mathbb{E}\left[ A^\pi(s, a)\middle|a\sim \pi'(.|O(s)) \right]\\
 & =\frac{1}{1-\gamma } \ \sum _{s\in S'}\sum _{o\in O} p^{\pi'}( s|o) p^{\pi'}( o)\mathbb{E}\left[ A^\pi(s, a)\middle|a\sim \pi'(.|O(s)) \right]\\
 &=\frac{1}{1-\gamma } \ \sum _{o\in O}p^{\pi'}( o)\sum _{s\in S'} p( s|o) \mathbb{E}\left[ A^\pi(s, a)\middle|a\sim \pi'(.|o) \right] \label{step:a}\\
 & =\frac{1}{1-\gamma } \ \sum _{o\in O} p^{\pi'}( o)\mathbb{E}\left[ A^\pi(o, a)\middle|a\sim \pi'(.|o)\right]
\end{align}
In Eq<~\eqref{step:a}: i) we can replace $O(s)$ by $o$ because $p( s|o)\neq 0$ only when $O(s) = o$. ii) The dependence on $\pi'$ for $p^{\pi'}(s|o)$ is dropped. We reiterate that this may not hold in general, but does when when the base task is a supervised learning task because base actions (e.g. predicting a class label for an input) bears no effect on the next base state distribution. 

\subsection{Theorem~\ref{thm:rpiconv}: convergence of Entropy Regularized Policy Iteration (ERPI)}
Let $\pi_1 : \Omega \mapsto \Delta\A'$ be a reactive policy uniform over the action space, $\pi_1(a|s) = \frac{1}{|\A'|}$. At each iteration $k$, ERPI learns first the Q-function $Q^{\pi_k}$, which we more simply denote $Q_k$ in the following, and then obtains policy $\pi_{k+1}$ by the following optimization for every $o \in \Omega$ 
\begin{equation}
\pi _{k+1}( .|o) \ =\arg\min_{p} \ \langle p,\ -Q_{k}( o,\ .) \rangle \ +\ \frac{1}{\eta }\text{KL}( p( .|o) \ |\pi _{k}( .|o)) ,
\end{equation}
where $\langle, \rangle$ denotes the dot product, $\displaystyle Q_{k}( o,\ a) \ =\ \sum _{s\in S} p( s|o) Q_{k}( s,\ a)$, and KL is the Kullback-Leibler divergence. As a result we have, $\pi _{k+1} \propto \pi _{k}\exp( \eta Q_{k})$, which implies $\pi _{k+1} \propto \exp( \eta \sum_{t=1}^{k}Q_{t})$. We previously described ERPI as a sum of past advantage functions but we note that this describes the same policy since $A_k(o, a) = Q_k(o, a) - V_k(o)$ and $V_k(o)$ does not depend on $a$.

\textbf{For a given $o \in \Omega$} (we drop dependence on $\displaystyle o$ for $\displaystyle \pi $ and $\displaystyle Q$ for clarity) we let $\displaystyle J_{k}( \pi ) \ =\ \langle \pi ,-Q_{k} \rangle +\frac{1}{\eta }\text{KL}( \pi |\pi _{k})$. $\displaystyle J_{k}$ is convex and for policy $\displaystyle \pi _{k+1}$ minimizing $\displaystyle J_{k}$, we have from the optimality condition of convex functions
\begin{align}
\label{eq:optimcond}
\langle \nabla J_{k}( \pi _{k+1}) ,\pi ^{*} \ -\pi _{k+1} \rangle \ \geq 0.\ \
\end{align}
We also have that
\begin{align}
\langle \nabla J_{k}( \pi _{k+1}) ,\pi ^{*} \ -\pi _{k+1} \rangle  & =\ \langle -Q_{k} +\frac{1}{\eta } \nabla \text{KL}( \pi _{k+1} |\pi _{k}) ,\ \pi ^{*} -\pi _{k+1} \rangle, \\
 & =\langle -Q_{k} ,\ \pi ^{*} -\pi _{k+1} \rangle +\frac{1}{\eta } \langle \nabla \text{KL}( \pi _{k+1} |\pi _{k}) ,\ \pi ^{*} -\pi _{k+1} \rangle, \\
 & =\langle -Q_{k} ,\ \pi ^{*} -\pi _{k+1} \rangle +\frac{1}{\eta }\left(\text{KL}\left( \pi ^{*} |\pi _{k}\right) -\text{KL}\left( \pi ^{*} |\pi _{k+1}\right) -\text{KL}( \pi _{k+1} |\pi _{k})\right),\label{step:b}
\end{align}
where Eq.~\eqref{step:b} follows from the generalized triangle equality of Bregman divergences \cite{Bubeck15}. Combining Eq.~\eqref{step:b} and Eq.\eqref{eq:optimcond} yields
\begin{align}
\label{eq:c}
\langle \pi _{k+1} ,-Q_{k} \ \rangle -\langle \pi ^{*} ,-Q_{k} \ \rangle  & \leq \frac{1}{\eta }\left(\text{KL}\left( \pi ^{*} |\pi _{k}\right) -\text{KL}\left( \pi ^{*} |\pi _{k+1}\right) -\text{KL}( \pi _{k+1} |\pi _{k})\right).
\end{align}
We also have that
\begin{align*}
J_{k}( \pi _{k}) & =\langle \pi _{k} ,-Q_{k} \ \rangle.
\end{align*}
Adding $\displaystyle J_{k}( \pi _{k}) -\langle \pi _{k+1} ,-Q_{k} \ \rangle $ to both sides of yields \ref{eq:c}

\begin{align}
\langle \pi _{k} ,-Q_{k} \ \rangle -\langle \pi ^{*} ,-Q_{k} \ \rangle  & \leq \frac{1}{\eta }\left(\text{KL}\left( \pi ^{*} |\pi _{k}\right) -\text{KL}\left( \pi ^{*} |\pi _{k+1}\right)\right) +J_{k}( \pi _{k}) -J_{k}( \pi _{k+1}). \label{eq:e}
\end{align}
\textbf{Upper bounding }$\displaystyle J_{k}( \pi _{k}) -J_{k}( \pi _{k+1})$:
We know that $\displaystyle \pi _{k+1} =\frac{\pi _{k}\exp( \eta Q_{k})}{Z}$ with $\displaystyle Z=\langle \pi _{k} ,\exp( \eta Q_{k}) \rangle $. The KL-divergence becomes then

\begin{align}
\text{KL}( \pi _{k+1} \mid \pi _{k}) & =\langle \pi _{k+1} ,\log\frac{\pi _{k}\exp( \eta Q_{k})}{Z\pi _{k}} \rangle, \\
 & =\langle \pi _{k+1} ,\log\frac{\exp( \eta Q_{k})}{Z} \rangle, \\
 & =\eta \langle \pi _{k+1} ,Q_{k} \rangle -\log Z.
\end{align}
Using this results with the definition of $J_k$ gives
\begin{align}
J_{k}( \pi _{k}) -J_{k}( \pi _{k+1}) & =\langle \pi _{k} ,-Q_{k} \ \rangle -\langle \pi _{k+1} ,-Q_{k} \rangle -\frac{1}{\eta }( \eta \langle \pi _{k+1} ,Q_{k} \rangle -\log Z)\\
 & =\langle \pi _{k} ,-Q_{k} \ \rangle +\frac{1}{\eta }\log \ \langle \pi _{k} ,\exp( \eta Q_{k}) \rangle \\
 & \leq \langle \pi _{k} ,-Q_{k} \ \rangle +\frac{1}{\eta }\left( \eta \langle \pi _{k} ,\ Q_{k} \rangle \ +\ \frac{\eta ^{2}( Q_{\max} -Q_{\min})^{2}}{8}\right)\label{eq:d}\\
 & =\eta \frac{( Q_{\max} -Q_{\min})^{2}}{8}\\
 & \dot{=} \eta C
\end{align}
Inequality \eqref{eq:d} uses Hoeffding's lemma. Using this last result in Eq.~\eqref{eq:e} and averaging over K iterations yields
\begin{align}
\frac{1}{K}\sum _{k=1}^{K} \langle \pi ^{*} ,Q_{k} \ \rangle -\langle \pi _{k} ,Q_{k} \ \rangle  & \leq \frac{1}{\eta K}\sum _{k=1}^{K}\text{KL}\left( \pi ^{*} |\pi _{k}\right) -\text{KL}\left( \pi ^{*} |\pi _{k+1}\right) +\eta C\\
 & \leq \frac{1}{\eta K}\text{KL}\left( \pi ^{*} |\pi _{1}\right) +\eta C\\
 & =\frac{1}{\eta K} \langle \pi ^{*} ,\ \log |\A'|\pi ^{*} \rangle +\eta C\\
 & \leq \frac{1}{\eta K}\log \langle \pi ^{*} ,\ |\A'|\pi ^{*} \rangle +\eta C\\
 & \leq \frac{1}{\eta K}\log |\A'|+\eta C \label{eq:g}
\end{align}

To synthesize, we have upper bounded, for any $o \in \Omega$,  $\frac{1}{K}\sum _{k=1}^{K} \langle \pi ^{*}(.|o) ,Q_{k}(o, .)\rangle -\langle \pi_{k}(.|o) ,Q_{k}(o, .)\rangle$. Lemma~\ref{lemma:pdl} applied to $\pi^*$ and $\pi_k$ states 
\begin{align}
J\left( \pi^*\right) -J ( \pi_k) & =\frac{1}{1-\gamma } \ \sum _{o\in O} p^{\pi^*}( o)\mathbb{E}\left[ A^\pi(o, a)\middle|a\sim \pi'(.|o)\right]\\
&=\frac{1}{1-\gamma } \ \sum _{o\in O} p^{\pi^*}( o)(\langle\pi^*(.|o), Q_k(o, .)\rangle - \langle\pi_k(.|o), Q_k(o, .)\rangle\rangle)\label{eq:f}
\end{align}

Taking the expectation of \eqref{eq:g} over $\Omega$ with probability $\displaystyle p^{\pi^*}$, \ multiplying by $\displaystyle \frac{1}{1-\gamma }$ and using Eq.~\eqref{eq:f} yields

\begin{gather}
\frac{1}{K( 1-\gamma )}\sum _{o\in O} p^{\pi^*}( o)\sum _{k=1}^{K} \langle \pi ^{*}( .|o) ,Q_{k}( .|o) \rangle -\langle \pi _{k}( .|o) ,Q_{k}( .|o) \rangle \leq \frac{1}{1-\gamma }\left(\frac{1}{\eta K}\log |\A'|+\eta C\right)\\
\begin{aligned}
\Leftrightarrow \frac{1}{K}\sum _{k=1}^{K} J \left( \pi ^{*}\right) -J( \pi _{k}) & \leq \frac{1}{1-\gamma }\left(\frac{1}{\eta K}\log |\A'|+\eta C\right)\\
 & =\frac{1}{1-\gamma }\left(\frac{1}{\eta K}\log |\A'|+\eta \frac{( R_{\max} -R_{\min})^{2}}{8( 1-\gamma )^{2}}\right)\label{eq:h}
\end{aligned}
\end{gather}
Plugging in the optimal step-size $\displaystyle \eta =\frac{\sqrt{8\log |\A'|( 1-\gamma )}}{( R_{\max} -R_{\min})\sqrt{K}}$ in \eqref{eq:h} yields
\begin{align}
\frac{1}{K}\sum _{k=1}^{K} J \left( \pi ^{*}\right) -J ( \pi _{k}) & \leq \frac{1}{1-\gamma }\left(\frac{1}{\eta K}\log |\A'|+\eta \frac{( R_{\max} -R_{\min})^{2}}{8( 1-\gamma )^{2}}\right)\\
 & \leq \frac{\sqrt{2\log |\A'|} \ ( R_{\max} -R_{\min})}{2\sqrt{( 1-\gamma )^{3} K}}
\end{align}
Observing that $J \left( \pi ^{*}\right) -\max_{k\in\{1,\dots, K\}}J ( \pi_k) \leq \frac{1}{K}\sum _{k=1}^{K} J \left( \pi ^{*}\right) -J ( \pi _{k})$ completes the proof.
\subsection{Theorem~\ref{theo:obs}: problem equivalence with the Observation-IBMDP}
\label{ann:equi}
A reactive policy of an IBMDP $\pi: \Omega \mapsto \Delta\A'$ can act on the Observation-IBMDP since the state space of the latter MDP is $\Omega$ and its action space is $\A'$. We will show in this section that they also have the same policy return. Indeed, the construction of the Observation-IBMDP's transition function is such that feature bounds are visited with the same frequency in the IBMDP and the Observation-IBMDP, while the reward at state $o$ of the Observation-MDP is the average over the state rewards of the IBMDP that 'fall' within the feature bounds.

The proof holds only when the base MDP describes a supervised task because again we have that for any reactive policy acting in the IBMDP, $Pr(s_t = s|o_t = o) = p(s|o)$. That is, for a policy $\pi$ acting in the IBMDP, if at time-step $t$ the observation part of the state is $o$, then the distribution of the random variable $s_t$ follows the above fixed distribution. This distribution is simply the data distribution of features $\phi$ that 'fall' within $o$, i.e. for which each dimension of $\phi$ must be higher (resp. lower) than the lower (res. upper) bound described by $o$. 

Let $o_t$ and $v_t$, $t\geq0$ be the feature bounds random variables as $\pi$ acts on the IBMDP and the Observation-IBMDP respectively. We will show that for all $o\in\Omega$ and $t\geq 0, Pr(o_t = o) = Pr(v_t = o)$. By induction, it is true for $t=0$ since the feature bounds are all initialized to $(0,1)$  for both the IBMDP and Observaion-IBMDP. Assume it is true for $t$ then 
\begin{align}
    Pr(o_{t+1} = o') &= \sum_{a\in\A'}\sum_{s, s'\in\SC'}Pr(s_t = s)\pi(a|O(s)) T'(s, a, s') \mathbbm{1}_{O(s') = o'}\\
    &= \sum_{o\in\Omega}\sum_{a\in\A'}\sum_{s, s'\in\SC'}Pr(s_t = s|o_t = o)Pr(o_t=o)\pi(a|O(s)) T'(s, a, s') \mathbbm{1}_{O(s') = o'}\\
    &= \sum_{o\in\Omega}Pr(o_t=o)\sum_{a\in\A'}\pi(a|o)\sum_{s, s'\in\SC'}Pr(s_t = s|o_t = o) T'(s, a, s') \mathbbm{1}_{O(s') = o'}\\
    &= \sum_{o\in\Omega}Pr(o_t=o)\sum_{a\in\A'}\pi(a|o)\sum_{s, s'\in\SC'}p(s|o) T'(s, a, s') \mathbbm{1}_{O(s') = o'}\\
    &= \sum_{o\in\Omega}Pr(v_t=o)\sum_{a\in\A'}\pi(a|o)\sum_{s, s'\in\SC'}T''(o, a, o')\\
    &=Pr(v_{t+1} = o')
\end{align}
Now let $J_O(\pi)$ be the policy return of $\pi$ when acting on the Observation-IBMDP. We have 
\begin{align}
J(\pi) &= \sum_{t\geq0}\gamma^t \sum_{s\in\SC'}\sum_{a\in\A'}Pr(s_t=s)\pi(a|O(s))R(s, a)\\
&= \sum_{t\geq0}\gamma^t \sum_{s\in\SC'}\sum_{o\in\Omega}\sum_{a\in\A'}Pr(s_t=s|o_t=o)Pr(o_t=o)\pi(a|O(s))R'(s, a)\\
&=\sum_{t\geq0} \gamma^t\sum_{o\in\Omega}\sum_{a\in\A'}Pr(v_t=o)\pi(a|o)\sum_{s\in\SC'}p(s|o)R'(s, a)\\
&=\sum_{t\geq0} \gamma^t\sum_{o\in\Omega}\sum_{a\in\A'}Pr(v_t=o)\pi(a|o)R''(o, a)\\
&= J_{O}(\pi)
\end{align}
Thus reactive policies have the same policy return in the IBMDP and the Observation-IBMDP and thus an optimal policy of the Observation-IBMDP has a return $J(\pi^*)$.

\subsection{Proposition~\ref{prop:deterministic}: Performance of greedy policy extracted from ERPI}
Let $\pi = \arg\max_{\pi\in\{\pi_1,\dots, \pi_K\}}J(\pi_k)$. Let $Q_O^\pi$ its Q-function in the Observation-MDP. Let $\pi'(o) = \arg\max_{a\in\A'}Q_O(o,a)$. By the policy improvement theorem \cite{sutton2018reinforcement} $J_O(\pi')\geq J_O(\pi)$, and from the results of the previous section \ref{ann:equi}, this implies that $J(\pi')\geq J(\pi)$ in the IBMDP which concludes the proof.

\section{IBMDPs for depth-2 balanced binary DTs}\label{appendix:toytask}
\subsection{Turning a supervised classification task into a MDP}\label{appendix:classif-to-mdp}
Any supervised classification task can be cast into a Markov decision problem. If the dataset to be classified is $\mathcal{X} \subsetneq \mathbb{R}^{N \times d}$ then the state space of the MDP is $\SC = \mathcal{X}$. If the number of classes is $K$, then the action space is $\A = \{C_1, ..., C_K\}$. The transition function is stochastic, we simply transition to a new state (draw a new data point to classifiy) whatever the action: $T(s, a, s') = \frac{1}{|\SC|}$. The reward function depends on the current state and action: $R(s, a) = R(x_n, C_k) = 1 \text{ if } x_n \in C_k \text{ in the base supervised classification task };R(s, a) = -1 \text{ otherwise}$. A policy $\pi: \SC \rightarrow \A$ is a classifier. And such policy $\pi$ that maximizes the expected discounted cumulative reward, also maximizes the classification accuracy. 
\subsection{IBMDPs of simple supervised classification tasks}\label{appendix:supibmdp}
\begin{figure}[!ht]
    \subfloat[]{%
      \includegraphics[width=0.22\textwidth]{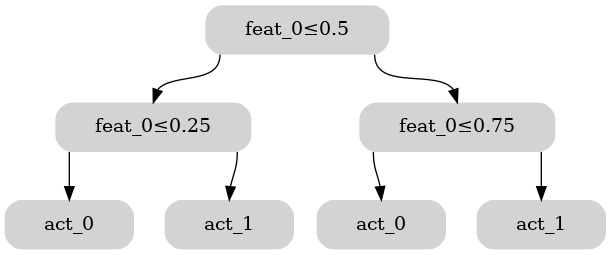}
    }
    \hfill
    \subfloat[]{%
      \includegraphics[width=0.22\textwidth]{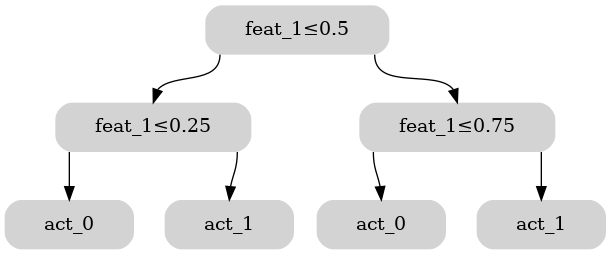}
    }
    \hfill
    \subfloat[]{%
      \includegraphics[width=0.22\textwidth]{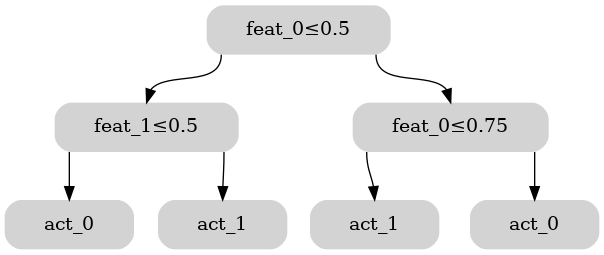}
    }
    \hfill
    \subfloat[]{%
      \includegraphics[width=0.22\textwidth]{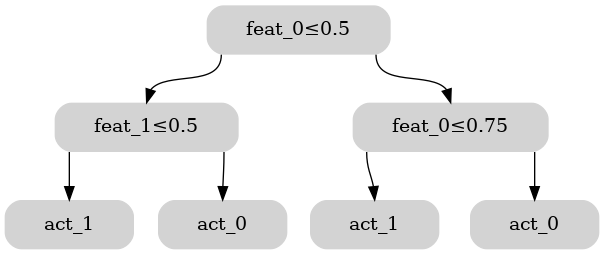}
    } \\
    \subfloat[]{%
      \includegraphics[width=0.3\textwidth]{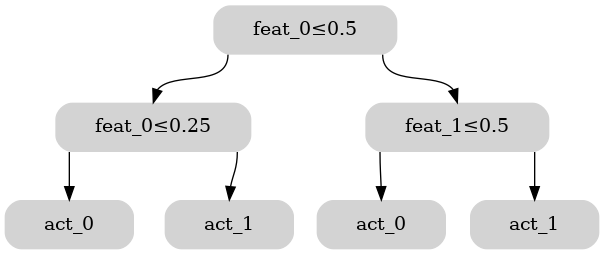}
    }
    \hfill
    \subfloat[]{%
      \includegraphics[width=0.3\textwidth]{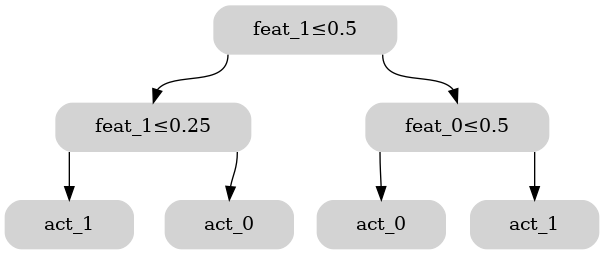}
    }
    \hfill
    \subfloat[]{%
      \includegraphics[width=0.3\textwidth]{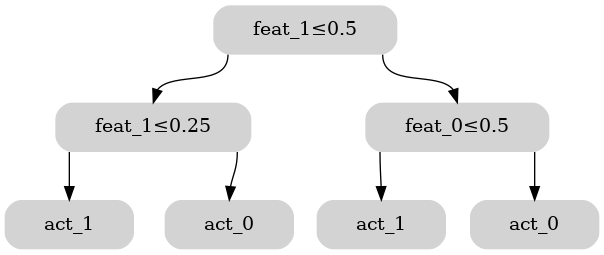}
    }
    \caption{Depth 2 binary Decision Trees}
    \label{fig:dt-ibmdp}
  \end{figure}

\begin{figure}[!ht]
    \centering
    \includegraphics[width=0.4\textwidth]{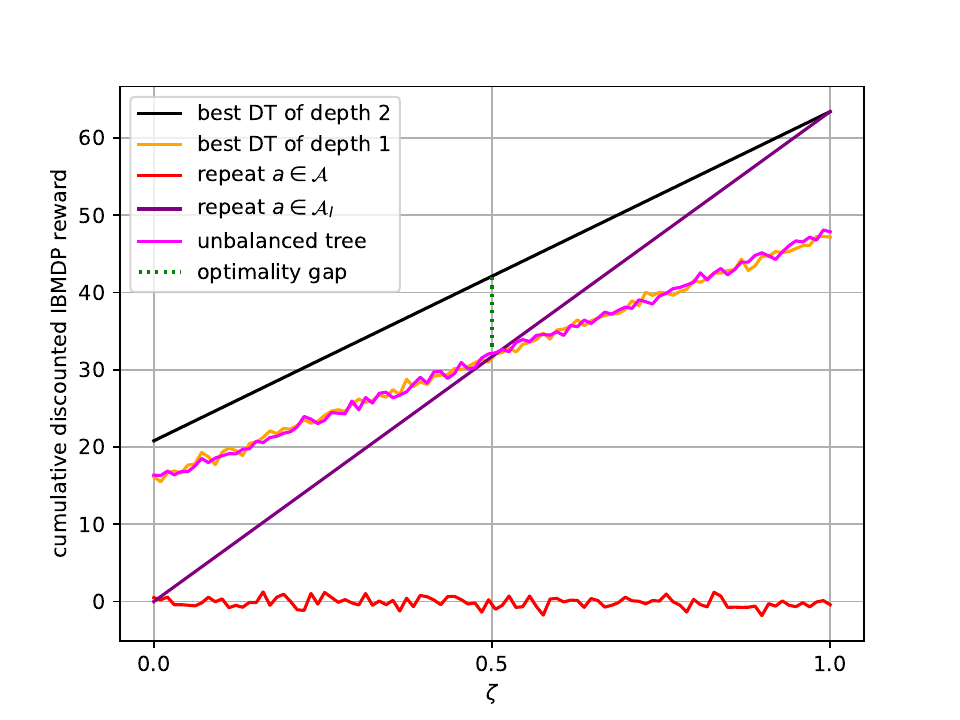}
    \caption{Cumulative discounted IBMDP reward of binary DTs as a function of $\zeta$}
    \label{fig:zeta-opt}
\end{figure}

We want to study the ability of algorithms to retrieve the optimal reactive policy in an IBMDP. To do so we design a set of seven base supervised binary classification tasks each made of 16 data points in $[0,1]^2$. Each of the seven tasks can be accurately solved by balanced binary DTs of depth 2 (the DTs are presented on Fig. \ref{fig:dt-ibmdp}). 
\paragraph{Supervised classification MDP:} 
We definethe MDP associated with those supervised binary classification tasks as in Sec. \ref{appendix:classif-to-mdp}. There are as many states as there are data points to classify. In particular, the states are $\SC = \{0.125, 0.375, 0.625, 0.875\}^2$. There are two actions, one for each classes in the task: $\A = \{C_1, C_2\}$. The reward function depends on the current state and action: $R(s, a) = 1 \text{ if } s \in a;R(s, a) = -1 \text{ otherwise}$. The transition function is stochastic, we simply transition to a new state (draw a new data point to classify) whatever the action: $T(s, a, s') = \frac{1}{|\SC|}$. Finally there is a discount factor $\gamma$. 
\paragraph{Supervised classification IBMDP:} Now to find a policy (classifier) that is a DT for such supervised binary classification MDPs, we look for reactive policies in the following IBMDPs \cite{IBMDP} (see \ref{sec:ibmdp} for generalities about IBMDPs). Note that all trees in \ref{fig:dt-ibmdp} are of depth 2, and all split thresholds are in $\{0.25, 0.5, 0.75\}$, so we can have a finite state space for the IBMDPs (the observation part of the state space is finite, $o \in {(0,0,1,1), (0,0,0.75,1), (0,0,1,0.75), (0.25, 0, 1, 1), (0, 0.25, 1, 1), (0,0, 0.5, 1), ..., (0.5, 0.5, 1, 1) }$) and choose $p = 1$.
The IBMDP state space is the product product of the base classification MDP:
$\SC' \subsetneq \SC \times [0,1]^{2\times2}$. In particular, $\SC' = \{(0.125, 0.125, 0, 0, 1, 1), (0.125, 0.125, 0, 0, 0.5, 1), (0.125, 0.125, 0, 0, 1, 0.5), (0.125, 0.125, 0, 0, 0.5, 0.5) , $
$(0.375, 0.125, 0, 0, 1, 1), ..., (0.875, 0.875, 0, 0.75, 1, 1) \}$. 
$|\A'| = |\A| + |\A_I| = 2 + 2 = 4$ because there are two base actions ($\A = \{C_1, C_2\}$) and each feature bound can be split at $ p + 1 = 2$ thresholds. We choose $\zeta = 0.5$ so that the trees in Fig. \ref{fig:dt-ibmdp} are optimal with respect to the discounted cumulative reward of the IBMDPs. In Fig. \ref{fig:zeta-opt} we compare the value of the cumulative discounted IBMDP rewards obtained by the reactive policies that can be learned in the IBDMPs, including the one corresponding to the trees in Fig. \ref{fig:dt-ibmdp}. From Fig. \ref{fig:zeta-opt}. It is clear that choosing $\zeta = 0.5$ will facilitate the learning of reactive policies corresponding to Fig. \ref{fig:dt-ibmdp} by making them optimal w.r.t the cumulative discounted IBMDP rewards and maximizing the optimality gap with other reactive policies. 

\section{DTs obtained by ERPI, asymmetric PPO, and CART, for \texttt{wine} and \texttt{iris}}\label{appendix:trees}
\subsection{\texttt{iris}}
\begin{figure*}[!ht]
    \vskip -0.2in
    \subfloat[DT obtained by CART when $M = 1$, accuracy = 64\%]{
      \includegraphics[width=0.23\textwidth]{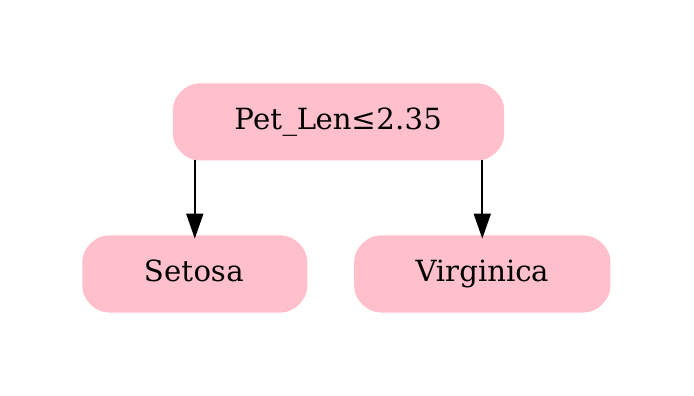}
    }
    \hfill
    \subfloat[DT obtained by CART when $M = 2$, accuracy = 93.3\%]{%
      \includegraphics[width=0.23\textwidth]{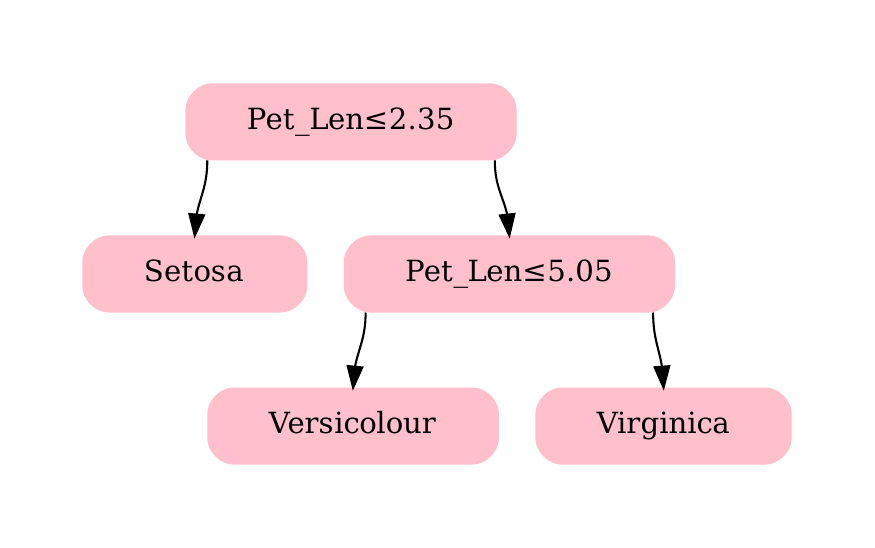}
    }
    \hfill
    \subfloat[DT obtained by CART when $M = 3$, accuracy = 94\%]{
      \includegraphics[width=0.23\textwidth]{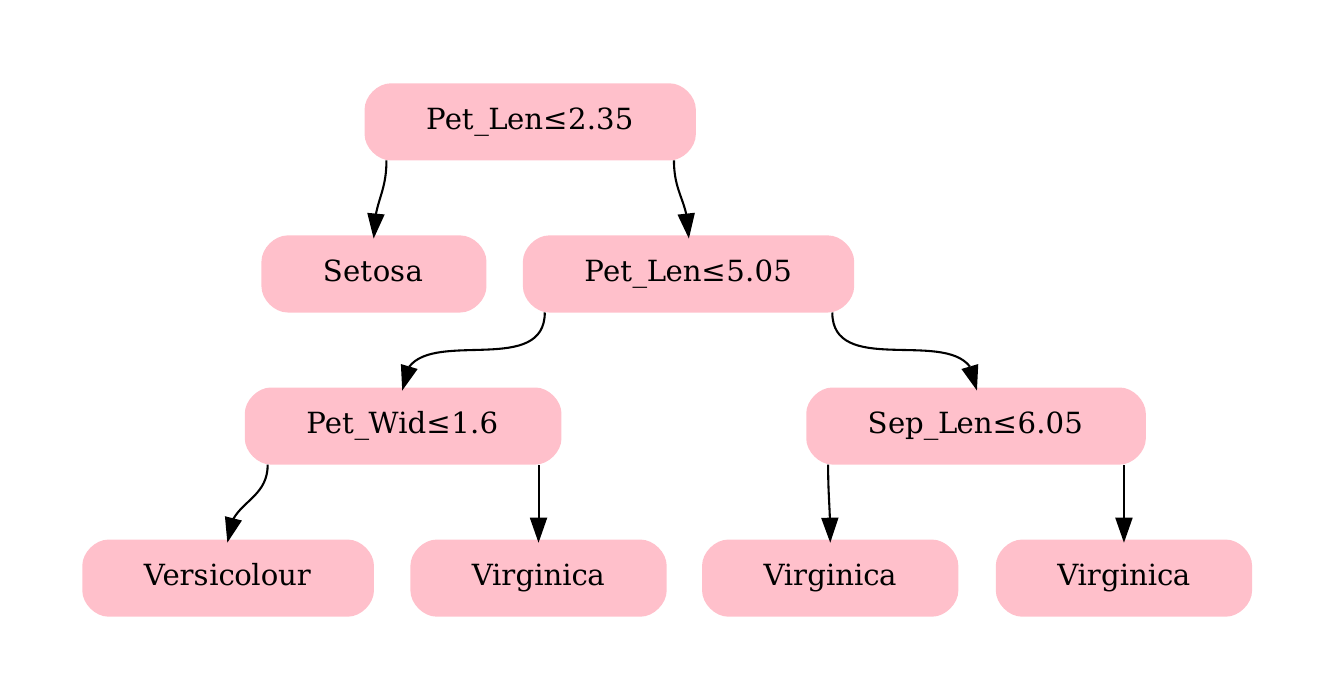}
    }
    \hfill
    \subfloat[DT obtained by CART when $M = 4$, , accuracy = 100\%]{%
      \includegraphics[width=0.23\textwidth]{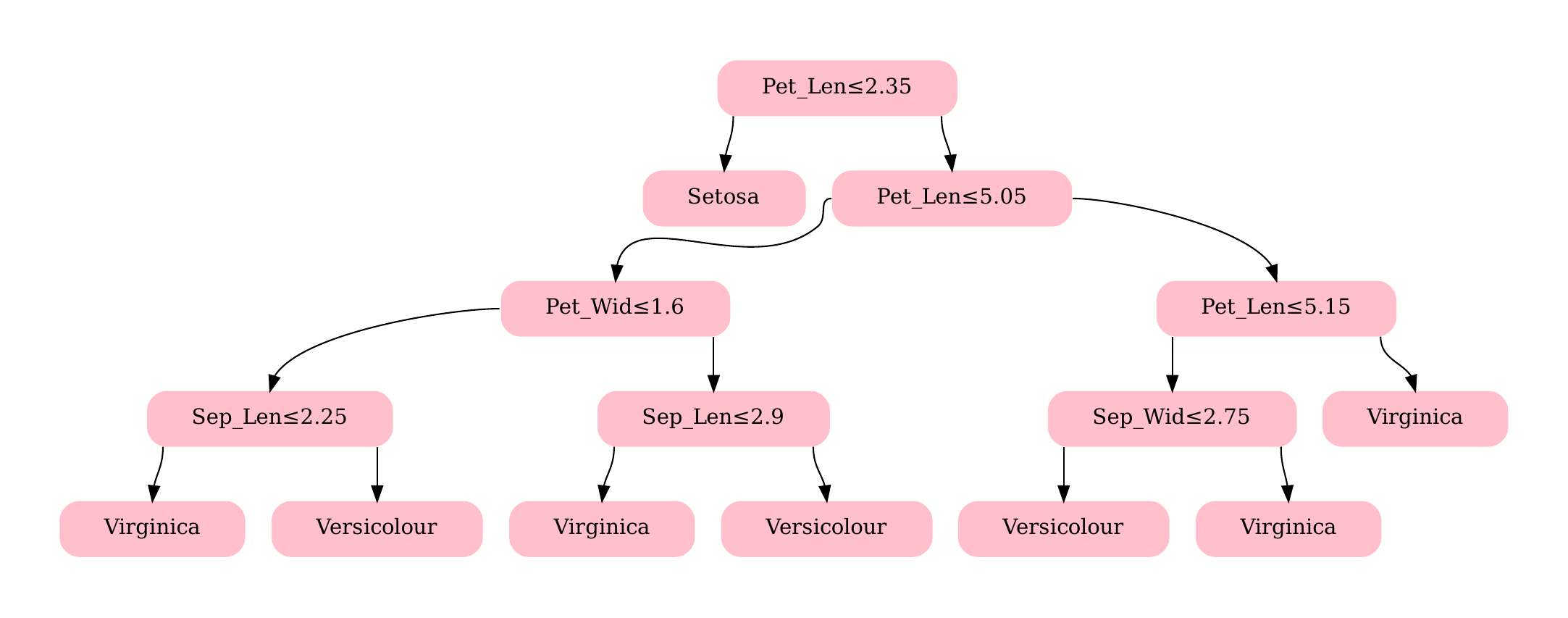}
    }
    \caption{DTs and their accuracy obtained by CART \texttt{iris} dataset.}
    \label{fig:iris-cart-trees}
  \end{figure*}

\begin{figure*}[!ht]
    \vskip -0.2in
    \subfloat[DT obtained by ERPI when $\zeta = -0.8$, accuracy = 34\%]{
      \includegraphics[width=0.23\textwidth]{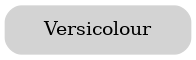}
    }
    \hfill
    \subfloat[DT obtained by ERPI when $\zeta = 0$, accuracy = 64\%]{%
      \includegraphics[width=0.23\textwidth]{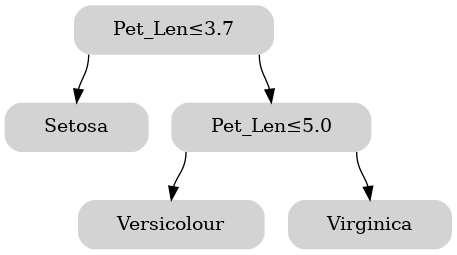}
    }
    \hfill
    \subfloat[DT obtained by ERPI when $\zeta = 0.8$, accuracy = 93.3\%]{
      \includegraphics[width=0.23\textwidth]{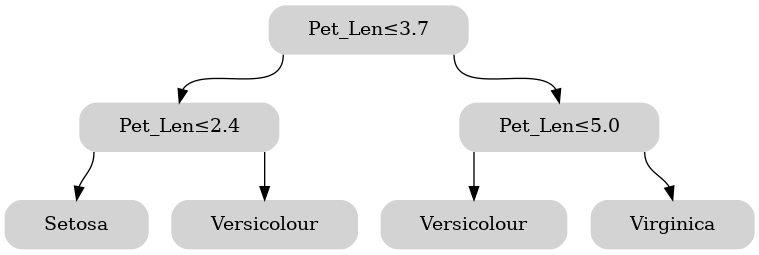}
    }

    \caption{DTs and their accuracy obtained by ERPI \texttt{iris} dataset.}
    \label{fig:iris-erpi-trees}
  \end{figure*}

\begin{figure*}[!ht]
    \vskip -0.3in
    \hfill
    \subfloat[DT obtained by asym PPO when $\zeta = -0.8$, accuracy = 34\%]{
      \includegraphics[width=0.23\textwidth]{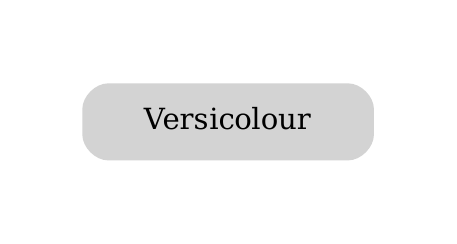}
    }
    \hfill
    \subfloat[DT obtained by asym PPO when $\zeta = 0$, accuracy = 64\%]{%
      \includegraphics[width=0.23\textwidth]{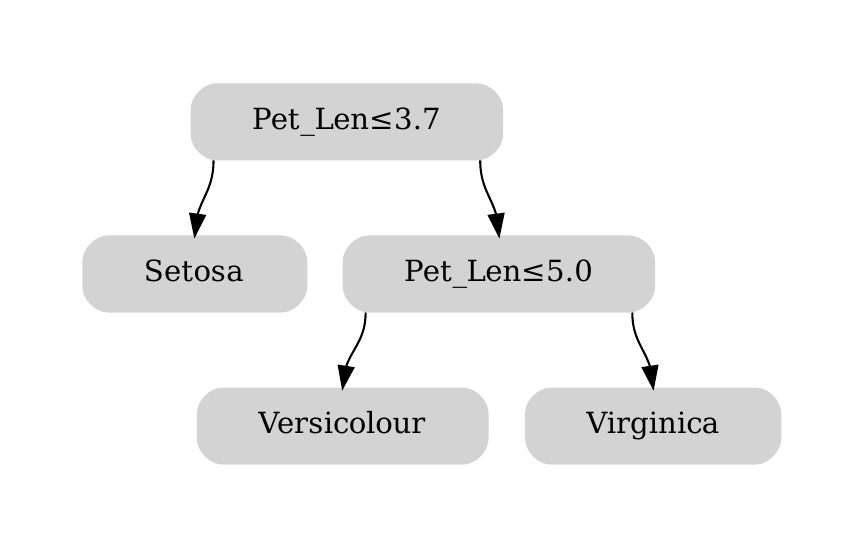}
    }
    
    \caption{DTs and their accuracy obtained by asymmetric PPO \texttt{iris} dataset.}
    \label{fig:iris-ppo-trees}
  \end{figure*}
\newpage
\subsection{\texttt{wine}}
\begin{figure*}[!ht]
    \vskip -0.3in
    \hfill
    \subfloat[DT obtained by CART when $M=1$, accuracy = 69\%]{
      \includegraphics[width=0.23\textwidth]{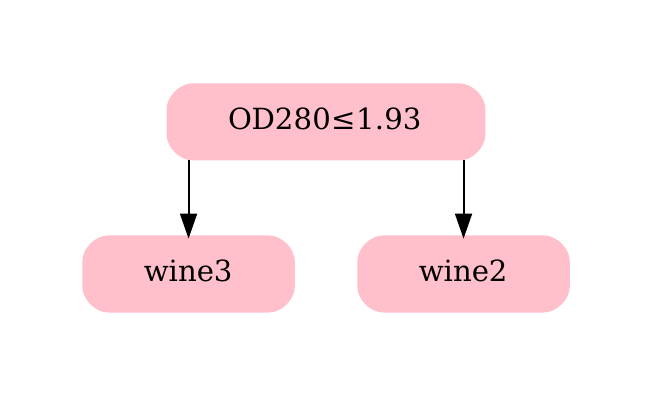}
    }
    \hfill
    \subfloat[DT obtained by CART when $M=2$, accuracy = 97\%]{%
      \includegraphics[width=0.23\textwidth]{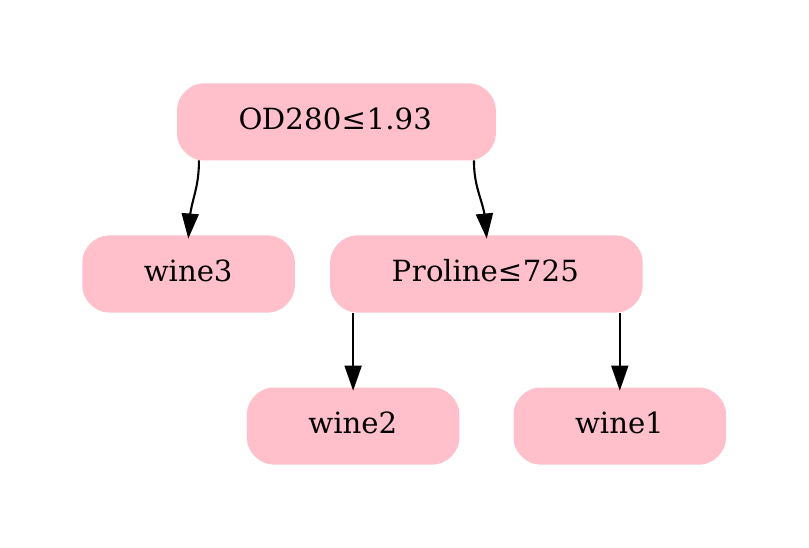}
    }
    \caption{DTs and their accuracy obtained by CART \texttt{wine} dataset.}
    \label{fig:wine-cart-trees}
  \end{figure*}

\begin{figure*}[!ht]
    \vskip -0.3in
    \subfloat[DT obtained by ERPI when $\zeta = -1$, accuracy = 34\%]{
      \includegraphics[width=0.23\textwidth]{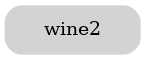}
    }
    \hfill
    \subfloat[DT obtained by ERPI when $\zeta = -0.6$, accuracy = 84\%]{%
      \includegraphics[width=0.23\textwidth]{plots/trees_wine/exact_computed_policy_zeta-0.6_eta1.png
    }}
    \hfill
    \subfloat[DT obtained by ERPI when $\zeta = 0.8$, accuracy = 95\%]{
      \includegraphics[width=0.23\textwidth]{plots/trees_wine/exact_computed_policy_zeta0.8_eta1.png}
    }

    \caption{DTs and their accuracy obtained by ERPI \texttt{wine} dataset.}
    \label{fig:wine-erpi-trees}
  \end{figure*}

\begin{figure*}[!ht]
    \vskip -0.7in
    \hfill
    \subfloat[DT obtained by asym PPO when $\zeta = -1$, accuracy = 34\%]{
      \includegraphics[width=0.23\textwidth]{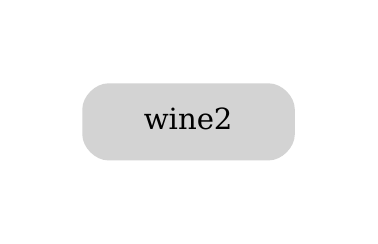}
    }
    \hfill
    \subfloat[DT obtained by asym PPO when $\zeta = -0.6$, accuracy = 34\%]{%
      \includegraphics[width=0.23\textwidth]{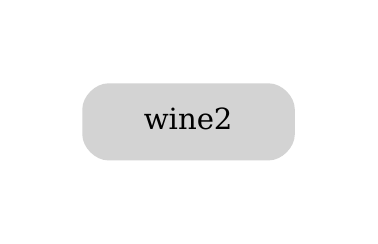}
    }
    
    \caption{DTs and their accuracy obtained by asymmetric PPO \texttt{wine} dataset.}
    \label{fig:wine-ppo-trees}
  \end{figure*}

\end{document}